\documentclass{article}

     \PassOptionsToPackage{numbers, sort&compress}{natbib}
\usepackage[final]{neurips_arxiv}

\usepackage[utf8]{inputenc} % allow utf-8 input
\usepackage[T1]{fontenc}    % use 8-bit T1 fonts
\usepackage[pagebackref,breaklinks,colorlinks]{hyperref}
\usepackage{url}            % simple URL typesetting
\usepackage{booktabs}       % professional-quality tables
\usepackage{amsfonts}       % blackboard math symbols
\usepackage{nicefrac}       % compact symbols for 1/2, etc.
\usepackage{microtype}      % microtypography
\usepackage[table]{xcolor}
\usepackage{epigraph} 
\usepackage{graphicx}
\usepackage{color}
\usepackage{subfigure}
\usepackage{multirow}
\usepackage{diagbox}
\usepackage{url}
\usepackage{amsmath}
\usepackage{xspace}
\usepackage{amssymb}
\usepackage{pifont}% http://ctan.org/pkg/pifont

\definecolor{xmark}{HTML}{E62020}
\definecolor{checkmark}{HTML}{40826D}

\newcommand{\ours}{{SMART-840}\xspace}

\def\eg{\emph{e.g}.}
\def\ie{\emph{i.e}.}
\def\etc{\emph{etc}.}

\newcommand{\comment}[1]{}

\newcommand{\newcontent}[1]{#1}
\newcommand{\set}[1]{\left\{#1\right\}}

\usepackage{xparse}
\NewExpandableDocumentCommand\mx{O{c}m}
  {\multicolumn{2}{c|}{#1}}

\newcommand\mcnew[1]{\multicolumn{2}{c}{#1}}

\title{Evaluating Large Vision-and-Language Models on Children's Mathematical Olympiads}

\author{%
Anoop Cherian$^1$\quad Kuan-Chuan Peng$^1$\quad Suhas Lohit$^1$\quad Joanna Matthiesen$^2$\quad \\
\textbf{Kevin Smith}$^3$\quad  \textbf{Joshua B. Tenenbaum}$^3$\\
$^1$Mitsubishi Electric Research Labs, Cambridge, MA, $^2$Math Kangaroo USA NFP\\$^3$Massachusetts Institute of Technology, Cambribdge, MA\\
\href{https://smartdataset.github.io/smart840}{ https://smartdataset.github.io/smart840}
}

\begin{document}

\maketitle

\begin{abstract}
Recent years have seen a significant progress in the general-purpose problem solving abilities of large vision and language models (LVLMs), such as ChatGPT, Gemini, etc.; some of these breakthroughs even seem to enable AI models to outperform human abilities in varied tasks that demand higher-order cognitive skills. \emph{Are the current large AI models indeed capable of generalized problem solving as humans do?}  A systematic analysis of AI capabilities for joint vision and text reasoning, however, is missing in the current scientific literature. In this paper, we make an effort towards filling this gap, by evaluating state-of-the-art LVLMs on their mathematical and algorithmic reasoning abilities using visuo-linguistic problems from children's Olympiads. Specifically, we consider problems from the Mathematical Kangaroo (MK) Olympiad, which is a popular international competition targeted at children from grades 1-12, that tests children's deeper mathematical abilities using puzzles that are appropriately gauged to their age and skills. Using the puzzles from MK, we created a dataset, dubbed \emph{\ours}, consisting of 840 problems from years 2020-2024. With our dataset, we analyze LVLMs power on mathematical reasoning; their responses on our puzzles offer a direct way to compare against that of children. Our results show that modern LVLMs do demonstrate increasingly powerful reasoning skills in solving problems for higher grades, but lack the foundations to correctly answer problems designed for younger children. Further analysis shows that there is no significant correlation between the reasoning capabilities of AI models and that of young children, and their capabilities appear to be based on a different type of reasoning than the cumulative knowledge that underlies children's mathematics and logic skills.
\end{abstract}

\section{Introduction}
\epigraph{``\small{\emph{Mathematics is not about numbers, equations, computations, or algorithms: it is about understanding.}}''}{\textit{William Paul Thurston}}

Recent multimodal artificial intelligence frameworks incorporating large vision and language models (LVLMs), such as GPT-4o, DALL-E, Gemini, \etc, are seen to demonstrate outstanding reasoning capabilities~\cite{wei2022emergent}, seemingly flustering our established measures of machine intelligence~\cite{chollet2019measure,hernandez2017measure,legg2007collection,minsky1988society,lake2017building}.  These scaled up Transformer models~\cite{vaswani2017attention} trained on internet-scale datasets using purportedly simplistic training losses such as mask predictions, suddenly appear to have emergent abilities rivaling expert human intellect even on tasks demanding higher-level cognition. Such superior accomplishments naturally raises several questions: Are these models indeed capable of having core knowledge and generalizing it towards deriving innovative methods for problem solving? Are they equipped with the faculties to reason like children or are they exploiting implicit biases in their web-scale training datasets towards generating responses that are seemingly correct? Where do AI models rank in their generalized intellectual capacities against humans? 

There have been several recent studies that attempt to answer the above questions through novel datasets, tasks, and benchmarks, \eg, SMART-101~\cite{cherian2023deep}, M\textsc{ath}V\textsc{ista}~\cite{lu2024mathvista},  Math-Vision~\cite{wang2024measuring}, MathOdyssey~\cite{fang2024mathodyssey}, MathScape~\cite{zhou2024mathscape}, \etc ~While, all these datasets and tasks evaluate varied facets of the generative and reasoning abilities of LVLMs, they typically compare the performance of an LVLM against prior state-of-the-art (SOTA) AI models. While, some of these tasks even include human performances, these evaluations use relatively few human subjects, and do not include the diversity, demographics, background, and other subjective attributes that could influence the solution scheme, making the comparison of AI models to human performances to have significant room for speculation. Notably, there appears to be a lack of a systematic study that benchmarks the capabilities of current SOTA AI models against human cognition on the respective tasks at scale.

Contrary to current AI models that are potentially trained on web-scale data at once, humans develop their problem solving abilities over a period of development towards adulthood, and the type and nature of  problems that they can typically solve at different stages of their growth vary significantly. For example, a first grader may be able to solve problems related to tracing a given curved path, however a 12-th grader is expected to solve problems related to finding the intersection points of curves. On the one hand, this incremental nature of building knowledge is essential to the development of solid human problem solving~\cite{bjork2011making,ericsson1993role,hattie2007power}. On the other hand, this cumulative knowledge gathering also enforces an order to the way cognitive foundations are established in rational agents, \eg, a 12-th grader is implicitly assumed to have the knowledge to solve problems that a first grader may or may not be able to solve. If we want artificial generalist models that think and reason like intelligent humans, we should expect those models to reliably demonstrate more primitive concepts, in order to build up to reason about more complex problems.

Guided by this insight, we make a first attempt towards systematically comparing the performance of AI models against children's abilities over the period of their growth. Similar to  previous and contemporary studies, such as SMART-101~\cite{cherian2023deep}, M\textsc{ath}V\textsc{ista}~\cite{lu2024mathvista}, and Math-Vision~\cite{wang2024measuring}, we base our approach on the analysis of the problem-solving skills of LVLMs on mathematical and algorithmic reasoning problems selected from Mathematical Olympiads. In contrast to exams typically given in schools, that test the overall grasp of taught subject matter, Olympiads often incorporate problems that explore deeper understanding of concepts, critical thinking abilities, innovative ways of looking at data, and deriving connections across knowledge for solutions. Among many such math Olympiads  (such as IMO, AMC, \etc, that are aimed at higher grade students), one that we use to base our benchmark experiments in this paper is the international Mathematical Kangaroo (MK) Olympiad~\cite{kangaroo}, which is an international competition held since 1998 in 80 countries with nearly 45,000 student participants from the USA in 2024. Further, MK: i) offers competitions for all grades of children, consisting of age appropriate problems, thus allowing comparisons of the performance of AI models against children from varied age groups and skill levels, ii) uses a multiple choice answer format allowing for easy and objective evaluations that are directly comparable to children's responses, iii) do not demand the participants to have advanced mathematical background (\eg, calculus, analysis, \etc) for deriving the solutions, and iv) offers both text only and image-text problems, thus offering multimodal LVLM evaluation. These aspects make MK Olympiads suitable for our study, thus allowing for a clear gauging of AI's reasoning abilities  against the extent of developing human intellect.  

\begin{figure}
    \centering
    \includegraphics[width=14cm,trim={0cm 7.8cm 0cm 0.8cm},clip]{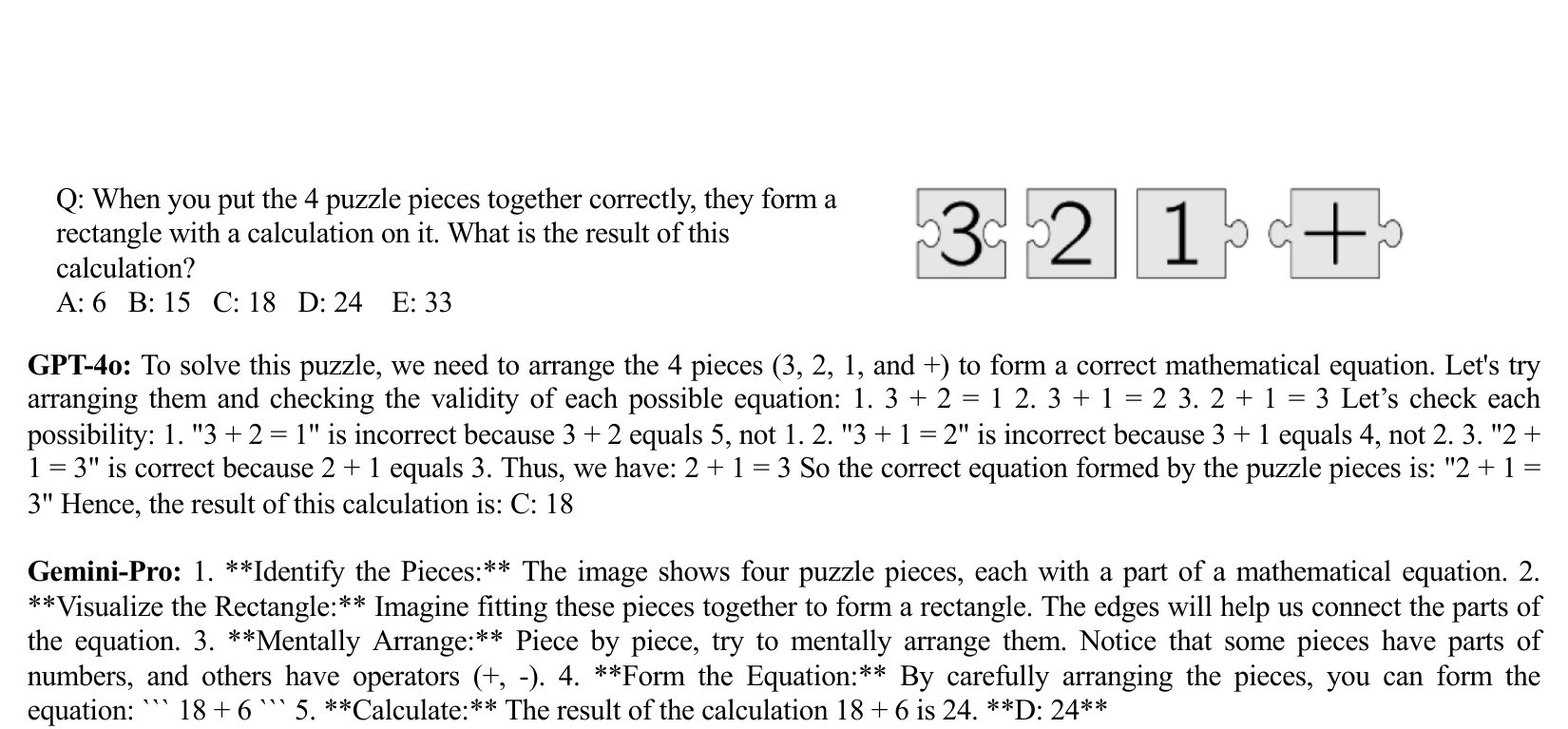}
    \caption{A 3rd grader puzzle from our \ours dataset and the LVLM responses (both incorrect).}
    \label{fig:teaser}
\end{figure}

To conduct our benchmarking experiments, we created an MK Olympiad dataset, dubbed \ours, by collating all the math competition problems from years 2000-2024 for all grades from 1-12th.  Our dataset offers a range of problem difficulties (some targeted at 1-2nd graders, some at 11-12th graders), and can further subdivide them by how children actually perform on those problems. Thus, for human-like mathematical reasoning we should expect: (a) better performance on earlier grade problems, and (b) a correlation with problem difficulty within grades. An example problem is provided in Figure~\ref{fig:teaser}. We also collect various performance statistics of children on all the exams from MK that are useful in our analysis, including the children's answer responses and time taken in deriving the solutions, among many others. Next, we shortlist several SOTA LVLMs, including closed-source models such as GPT-4o~\cite{openai2023gpt4}, Gemini~\cite{google2023geminipro}, and Claude-3~\cite{anthropic2023claudeopus}, and open-source models including LLaVa-Next~\cite{liu2023improvedllava}, and XGEN-MM-Phi3~\cite{xgen_mm_phi3_mini}, among others, that are popular on their generalist abilities.  Our analysis reveals several interesting aspects of the AI models: namely i) their performance on the \ours dataset appears to be significantly weaker than the average performance of children, ii) the performance gap is larger (nearly 30-40\%) on the tests for younger children and improves to 10-20\% for higher-schoolers, iii) there appears to be a lack of any significant correlation between the difficulty of a puzzle to children against that to an AI model, thus making it difficult to judge if a model could answer a given test problem, iv) there appears to be significant variability in the responses of current AI models when repeating or changing prompts, \newcontent{and v) recent LVLMs seem to demonstrate increasingly superior skills in solving \textit{text-only} problems, outperforming children}.  We believe our analysis brings new insights into various aspects for testing AI models against human cognition, as well as sheds light into the issues around reliability of current LVLMs for solving math problems.  

Before going further, we emphasize below the primary contributions of this paper.
\begin{enumerate}
    \item We provide a first of its kind benchmarking of the performance of large vision-and-language models against the mathematical and algorithmic reasoning abilities of young children using data from professional math Olympiads.
    \item We gauge the reasoning abilities of AI against the cumulative knowledge building progression of children over their growth.
    \item Our experiments compare SOTA AI models on both text-only and vision-and-text math problems,  analyzing the performances across multiple dimensions. 
\end{enumerate}

\section{Related Works}
\noindent  \textbf{General LVLM benchmarks}: Several benchmarks now exist that test different capabilities of LVLMs. These include MMBench~\cite{liu2023mmbench} which contains thousands of questions in a VQA format about both perception (1844 questions) and reasoning (1104) where the models select an answer from a given set of options. It also uses a ``circular evaluation'' strategy to ensure that the models are robust to the ordering of answer options. Although logical and relational reasoning are part of the dataset, this benchmark does not test particularly for different types of mathematical reasoning capabilities. MMMU~\cite{yue2023mmmu} is another popular benchmark that contains about 12.5K multimodal questions covering six different disciplines of study, but only at the college level and tests expert-level knowledge of LVLMs. In contrast, we are interested in understanding abilities of LVLMs that children demonstrate. A few benchmarks have been designed to test specific capabilities of LVLMs like ScienceQA ~\cite{lu2022learn} for scientific understanding and reasoning, VisIT-Bench~\cite{bitton2023visit} for instruction following, Bongard Problems ~\cite{bongard1968recognition, jiang2022bongard, jiang2022bhoi}, Raven's Progressive Matrices~\cite{benny2021scale}, Abstraction and Reasoning Corpus ~\cite{chollet2019measure} for abstract visual reasoning, OCRBench~\cite{liu2023hidden}, and TextVQA~\cite{singh2019towards} for text recognition, \etc  

\noindent \textbf{Benchmarks for mathematical reasoning}: M\textsc{ath}V\textsc{ista}~\cite{lu2024mathvista} is a recent benchmark for mathematical reasoning based on puzzles that involve images, while also measuring performance of LVLMs for different types of mathematical reasoning (logical, arithmetic, geometric, \etc) and different types of context images (natural images, line plots, scientific figures, \etc). GSM-8k~\cite{cobbe2021training} is a similar dataset containing about 8.5K math word problems, but only has text inputs and outputs, no images are involved. \newcontent{In GSM-Symbolic~\cite{mirzadeh2024gsm}, the fragility in logical reasoning of LLMs is explored}. The main difference of the proposed \ours benchmark against these prior works is that the these datasets neither separate puzzles based on hardness (\eg, ease of solving by children at different school levels), which our proposed benchmark explicitly addresses, nor are they supported by human performances at scale. TabMWP~\cite{lu2022dynamic} is a benchmark with about 38k grade-school problems but is limited to just tabular math word problems. SMART-101~\cite{cherian2023deep} is the most closely related benchmark to ours, that provides programmatically generated 2000 variations for each of 101 puzzles from just 1st and 2nd grade Math Kangaroo puzzles. These variations can be used to train larger models than using just a small number of puzzles. In contrast to this dataset, \ours contains 840 puzzles from all grades 1-12 and is designed to benchmark the zero-shot mathematical reasoning abilities of AI models on a wide range of problem solving skills, along with important information on performance statistics from test-takers. \newcontent{We note that we are not the first to use MK performance statistics for research, in fact in~\cite{mellroth2014high,mellroth2015problem,andritsch2020create,applebaum2019girls} MK exams were used to study the development of mathematical competencies in children, however to the best of our knowledge these tests have not been used to benchmark AI models.}

\section{Benchmarking Approach and Experiments}
We first elucidate our data collection process, followed by the details of the LVLMs that we select to benchmark in this study. The subsequent subsections evaluate the performances of LVLMs on various aspects of the Olympiads deriving correlations with the performances of children. 

\subsection{Mathematical Kangaroo Olympiad}
As alluded to above, most of the Mathematical Olympiads (such as International Math Olympiad, AMC-8, AHSME, \etc) are targeted at middle or high-school students, while Math Kangaroo is one Olympiad that conducts competitions for K-12 grades, making it a compelling source for this study. Started in France in 1991, the competition has been organized in the USA every year since 1998 and currently takes place in over 100 countries. Typically, there is a single exam for grades $\{n, n+1\}$, for $n\in\set{1, 3, 5, 7, 9, 11}$, thus there are a total of six exams in a year and children of both grades $n$ and $n+1$ participate in the same exam. 

Each exam consists of either 24 questions (for grades 1--4) or 30 questions for all higher grades, and is in a multiple choice format with 5 candidate answers, of which only one option is the correct answer. The questions can be purely text-based or can contain both text and an image, interpretation of both jointly is then usually important for solving the problem. Each question is attributed weights in $\set{3, 4, 5}$, where the lower points are given to problems that are typically deemed ``easier'' for that grade (\eg, single step reasoning problems for grade 1) while higher points are attributed to problems that need multi-level reasoning, enumeration of solution possibilities, \etc~that typically involve deeper (but age appropriate) problem solving skills. The participant is given 75 minutes to complete an exam and the performance is computed as the weighted sum of correct responses. 

\subsection{Data Collection}
\begin{figure}
    \centering
    \subfigure[]{\label{fig:participant_year_grade}\includegraphics[width=6cm]{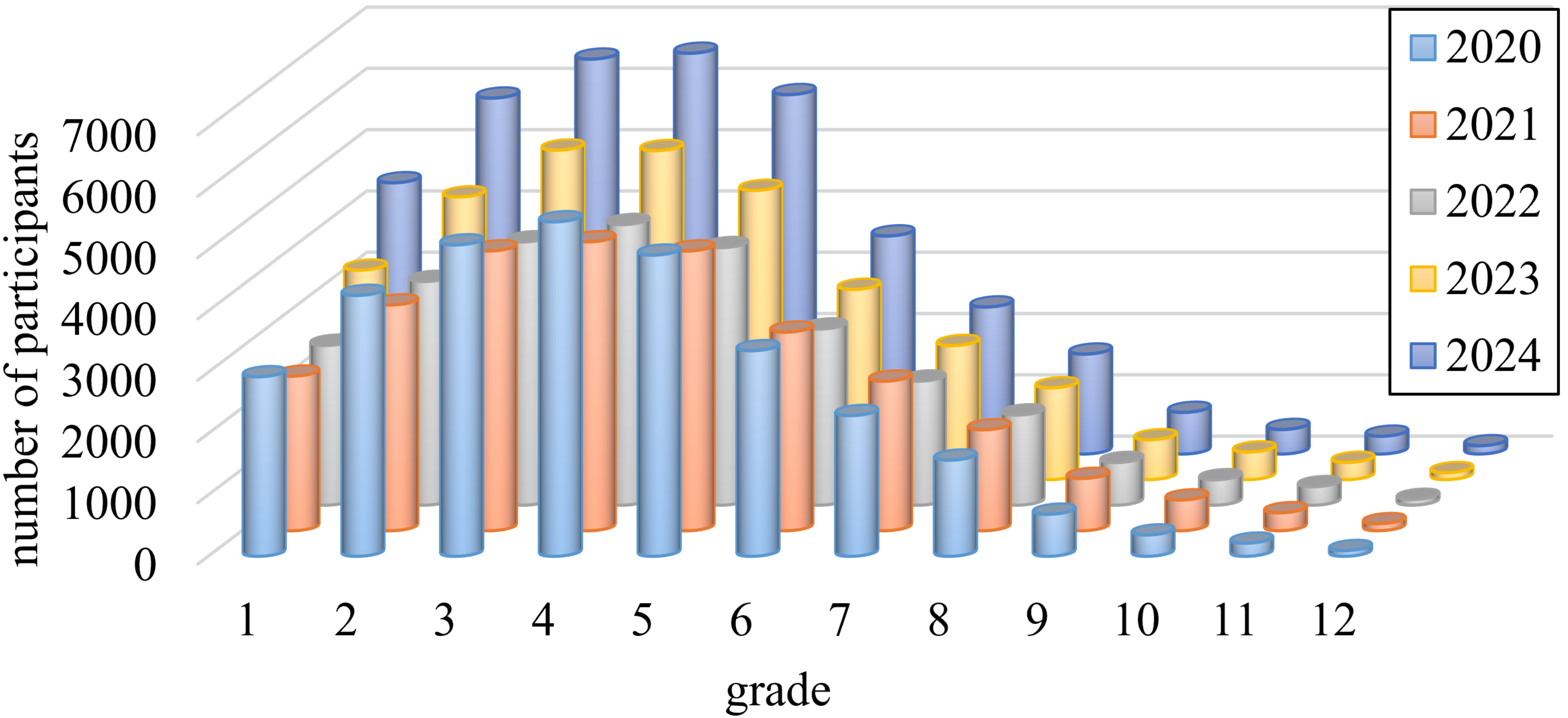}}
    \subfigure[]{\label{fig:participant_grade}\includegraphics[width=4.4cm]{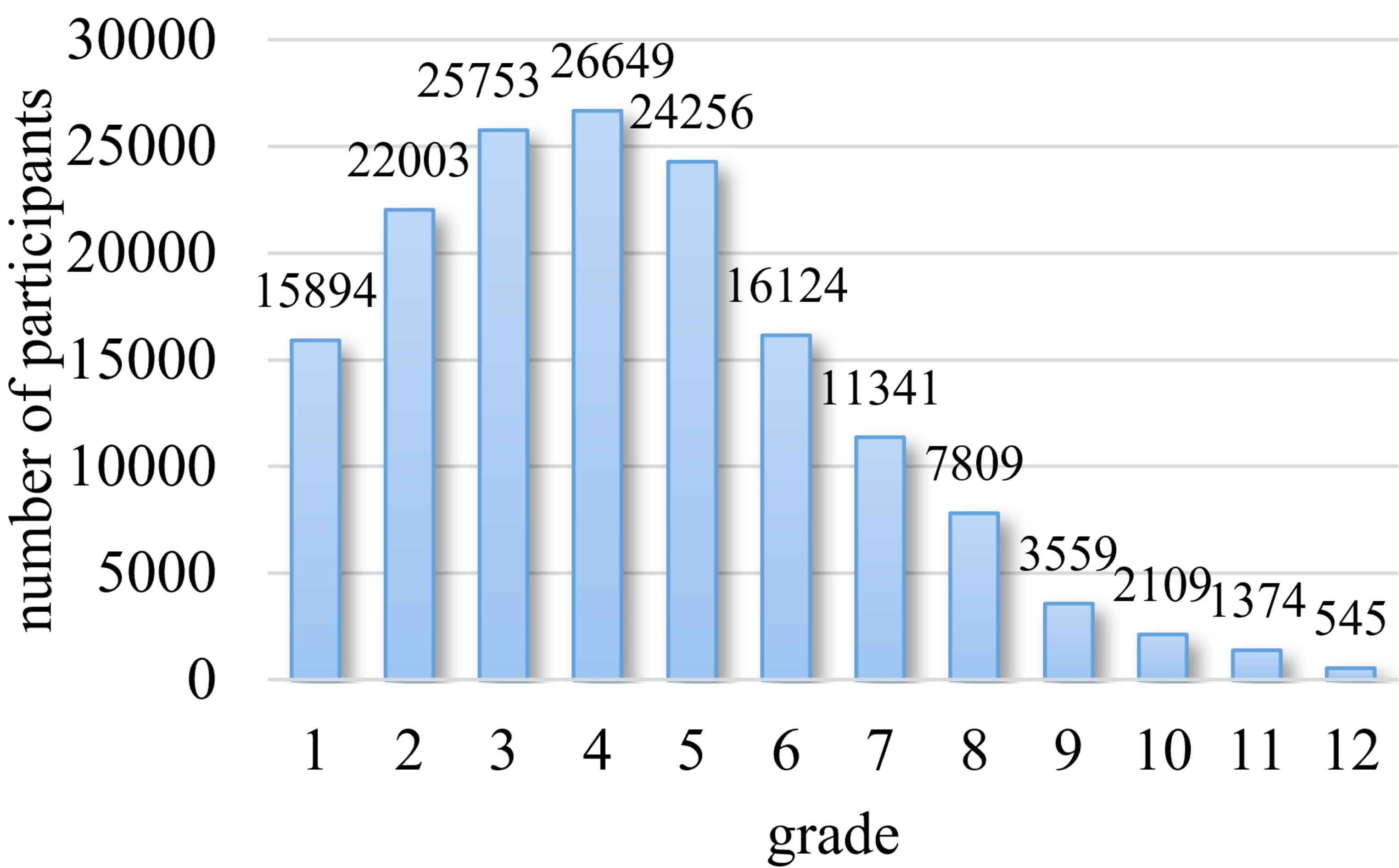}}
    \subfigure[]{\label{fig:participant_year}\includegraphics[width=3.4cm]{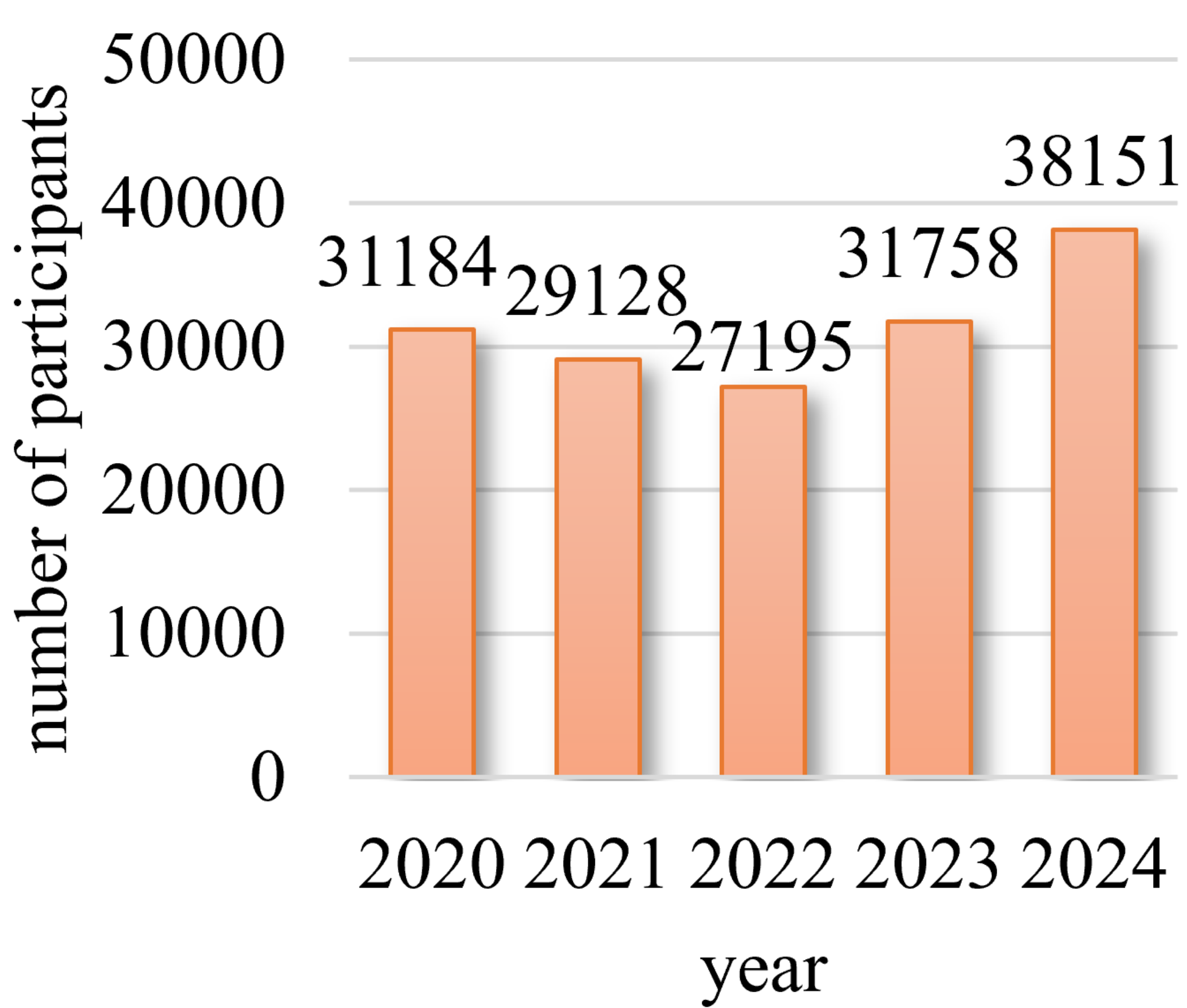}}
    \subfigure[]{\label{fig:pie_category}\includegraphics[width=4.6cm]{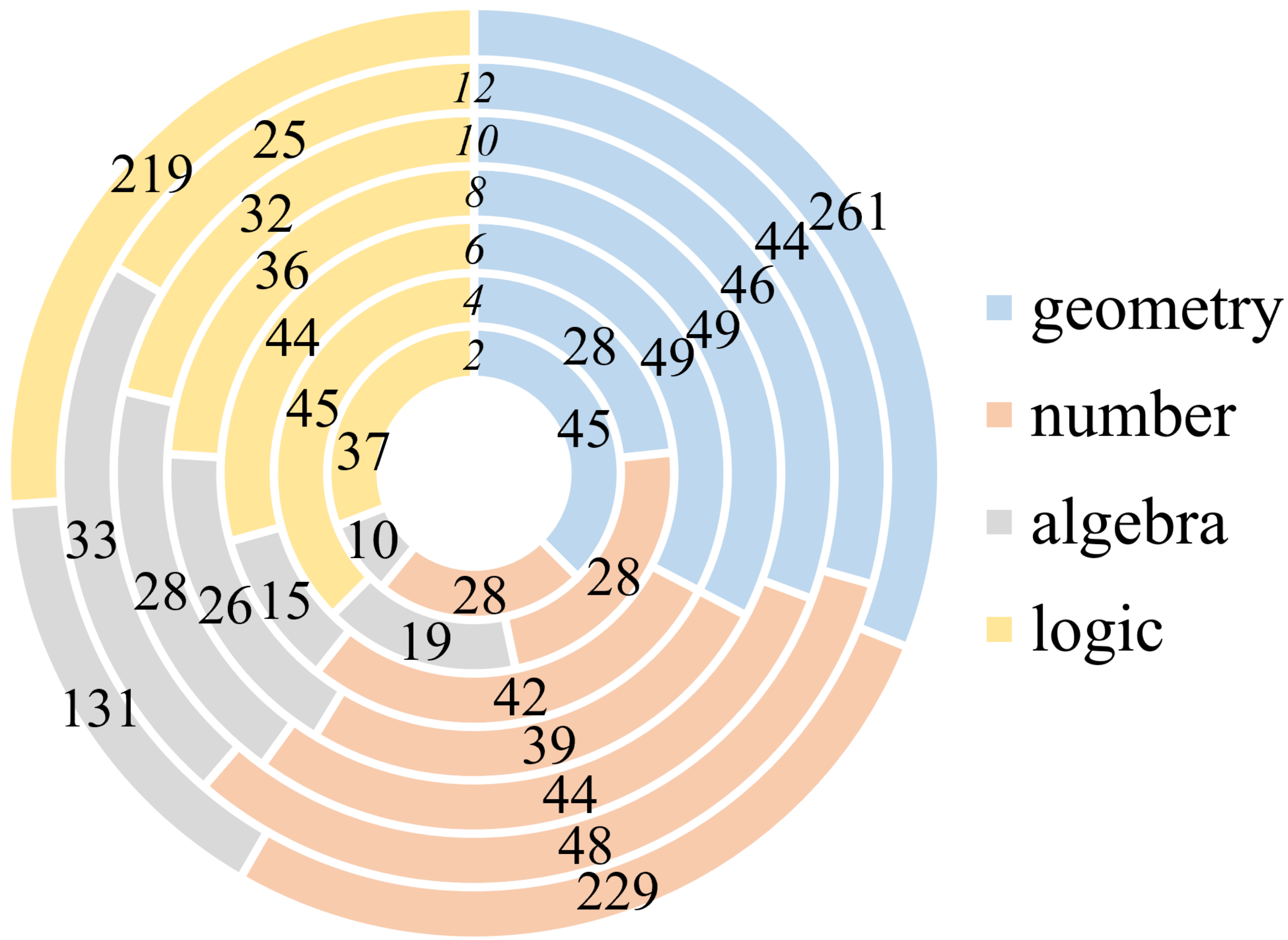}}
    \subfigure[]{\label{fig:pie_image_text}\includegraphics[width=4.6cm]{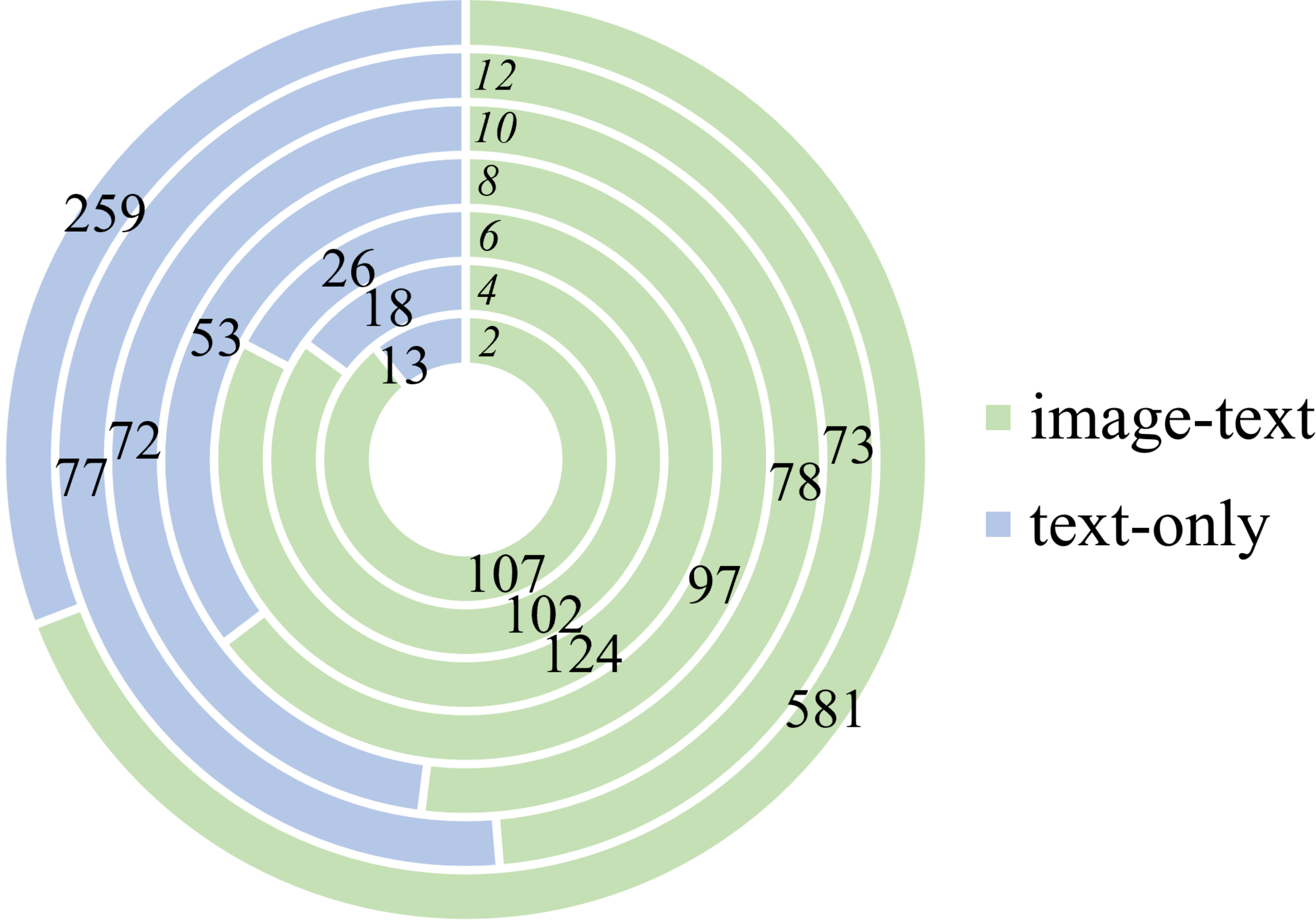}}
    \subfigure[]{\label{fig:pie_difficulty}\includegraphics[width=4.6cm]{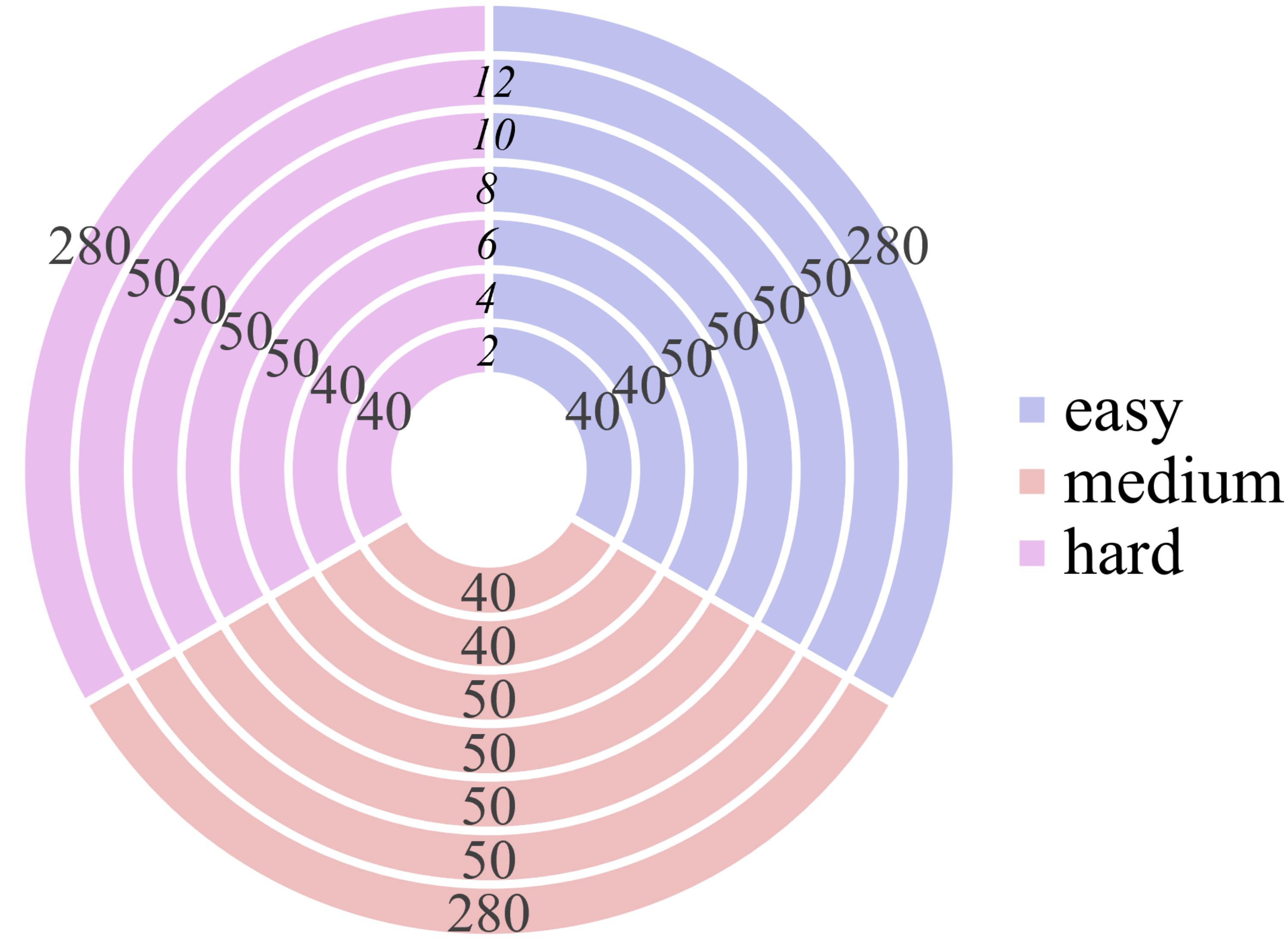}}
    
    \caption{Figure~\ref{fig:participant_year_grade} plots the distributions of children participating in MK Olympiads per year over 2020--2024 for grades 1--12. Figure~\ref{fig:participant_grade} plots the total number of participants per grade during 2020--2024. Figure~\ref{fig:participant_year} plots the total number of participants each year over all grades (1-12). Figure~\ref{fig:pie_category} shows the number of puzzles and its portion for each category. Figure~\ref{fig:pie_image_text} shows the statistics of image-text and text-only puzzles. Figure~\ref{fig:pie_difficulty} shows the statistics of puzzle difficulty (defined by their attributed weights). }
\end{figure}

For this study, we envisage a data collection methodology that is fair, balanced, and offers an unbiased benchmarking of AI models against children's performance on MK tests. To this end, we decided to use all the questions from MK exams without any omissions so that there is no selection bias in our evaluations. As the  statistical data we desire on the performance of children is unavailable for MK competitions prior to year 2020, in this study we consider only MK competitions from 2020-2024, that amounts to 840 problems in our dataset, dubbed \ours, and consisting of 240 questions all together from grades 1--4 and 600 questions from grades 5--12, evenly split between pairs of grades as described above. Figures~\ref{fig:participant_year_grade},~\ref{fig:participant_grade}, and~\ref{fig:participant_year} show the distribution of the number of children who participated across years from all grades, which adds to nearly 30K students per year. The participant number is highest for grades 1--8 and then drops to less than 1000 for grades 9--12 (Figure~\ref{fig:participant_grade}) perhaps because higher-grade children have other Olympiad options, \eg, AMC, IMO, \etc~Nevertheless, we see that the number of participants put together from the last five years still produce a substantially large sample set for our analysis.

For creating the \ours dataset, we downloaded publicly available\footnote{Note that each test paper involves a small cost for download.} question papers (which are image embedded PDF documents), followed by running optical character recognition software for extracting the text of the puzzles, and manually cropping the associated image parts. Each such extracted puzzle in the dataset was manually inspected for errors in its text and puzzle images. MK also provides a segregation of each puzzle into one of the four categories, namely (i) geometry, (ii) logic, (iii) algebra, and (iv) numbers. In Figure~\ref{fig:pie_category}, we present the overall statistics of problem distribution in the \ours dataset. We see that \emph{geometric} puzzles capture nearly 31\% of all the puzzles in our set, while the split is about equal between \emph{logic} (26\%) and \emph{numbers} (27\%), and \emph{algebra} based problems are about 15.5\%. In Figure~\ref{fig:pie_image_text}, we plot the distribution of the number of problems that need both text and image reasoning ($\sim$69\%) against those that only have text questions. Figures~\ref{fig:pie_category} and \ref{fig:pie_image_text} also show the split across grades. We see that for higher grades (>8), the number of text-only problems are higher: about 52\% in grades 11-12 against <20\% in grades 1-4.

\subsection{Selected Large Vision-and-Language Models}
We compare the performance of seven popular and SOTA LVLMs on the \ours dataset. Specifically, we consider i) GPT-4o~\cite{openai2023gpt4}, ii) Gemini-Pro~\cite{google2023geminipro}, and iii) Claude-3~\cite{anthropic2023claudeopus}, that are popular for their abilities in solving challenging math and visual reasoning problems. Thus, we believe it is a useful exercise to understand how they perform on children's grade problems. Alongside these SOTA LVLMs, we also consider other AI models that are popular, such as GPT-4v which is the first vision-and-language version of the GPT series, ii) Gemini-Flash that is well-known for its faster response time, and to recent open-source LVLMs such as \newcontent{XGen-MM-Phi3-Instruct-v1~\cite{xgen_mm_phi3_mini}, LlaVa-Next~\cite{liu2023improvedllava}, InternVL-Chat-V1-2 ~\cite{chen2024internvl}, and InternLM-XComposer-2.5~\cite{dong2024internlm}}. 

\subsection{Grade-wise Performance Comparisons}
In this experiment, we compare the performance of the LVLMs listed above against the performance of children on our \ours dataset. For the human performance, we report the percentage of average correct response rate, which we denote as \emph{accuracy} going forward, and is computed by: i) finding the ratio of the total number of correct children's responses on a problem to the total number of attempts, and (ii) averaging this ratio across all problems in the grade set. For the LVLMs, we use the API interface to query the model using a suitable hand-crafted prompt. Specifically, we found the following prompt to work well for all closed-source LVLMs: {\tt "Solve this question with explanation of the intermediate steps. Your response should end with one of the selected answer options from A1, B2, C3, D4, or E5."} which is accompanied by the text for the problem question and the image data.\footnote{The format of A1, B2, \etc~allow us to uniquely parse the LVLM response to automatically validate it. } For AI models, we report their accuracy as the (percentage) of problems correctly answered to the total number of problems in the set.  

In Table~\ref{tab:sota-comparisons}, we present results comparing the performances of LVLMs against children on the entire \ours dataset. First, we report a \emph{random} baseline to benchmark all our results, which is computed by randomly sampling a response from a probability distribution over all the human responses across the answer options for a problem. As is clear, all the answer options in the problems are equally likely, and thus the random performance is close to one-fifth. Next, for each LVLM, we queried (at least 2 times) each of the problems in \ours using the prompt described above. Note that for LVLM evaluation we also consider two additional possibilities, namely: (i) if a response is not in the expected format as demanded in the prompt, and if we are unable to automatically extract a valid response, we consider the response to be invalid in general (except in experiments when we manually validate the responses), and (ii) in many cases, an LVLM decides not to solve a problem (\eg, it mistakes the provided puzzle image to contain security issues), in which case as well, we declare that problem as unsolved by the respective model. We manually inspected all the output responses of GPT-4o (reported as GPT-4o (M) in the table) to ensure the our prompt is suitable, and the model produces responses that are reasonable, grounded in the problem specification (and are not due to  issues such as network failures, response parsing failures, etc.) and its solution attempt is reasonable (but not necessarily correct). All problems where the solution was unreasonable (even if the \emph{selected option} is correct), we manually marked them as a \emph{failed response}. 

\newcontent{We see from Table~\ref{tab:sota-comparisons} that GPT-4o demonstrates an accuracy of 42.5\% on average across all the grades, followed by Claude-3-Opus at 38\% and Gemini-Pro at nearly 32\%. The more recent Claude-3-Sonnet model in fact outperforms the performances of all the earlier models, with an average accuracy of 49.7\%, while outperforming GPT-4o in grades 1--6 and 9--10. There are several intriguing aspects in the performance of LVLMs that we can witness in Table~\ref{tab:sota-comparisons}.}

i) \textbf{Performance gap:} The performance of AI models are below that of children across the grades and interestingly this gap is consistent in all the models we experimented. Specifically, the best accuracy of LVLMs are in the range of 40-50\% while the children's average performance is consistently near 60\% or above. Note that we report children's performances for each grade separately, where kids of a pair of grades take the same exam. Unsurprisingly, we find that children of the higher grades perform significantly better than those of lower grades (although this gap reduces as problem solving abilities mature towards higher grades) suggesting a cumulative set of core problem solving skills that children build over their growth period.

ii) \textbf{{Performance trend:}} In Table~\ref{tab:sota-comparisons}, we see yet another consistent trend of LVLMs, \ie, being better at solving problems of higher grades (8-12) than at lower grades (\eg GPT-4o, Gemini-Pro, etc.) or similar performances in solving both higher- and lower-grader problems (\eg Claude-3 models), which is surprising given the complexity of solutions increases with grades. This trend was also seen in~\cite{cherian2023deep} where the authors compared the performance of LLMs on second grader problems. We see that while GPT-4o shows this increasing trend with an accuracy of 40\% at grades 1-2 towards nearly 50\% for grades 11-12, the trend is more striking for other LVLMs such as Gemini-Pro that varies by about 25\% (for grade 1-2) to 40\% for grades 11-12. We find that Claude-3 models produce a reasonably consistent performance around 40\% albeit having a different trend: dip in the performance for middle grades than lower or higher grades. 
 
\newcontent{We find from Table~\ref{tab:sota-comparisons} that open source LVLMs such as XGEN-MM, InternLM-XComposer2-VL and InternVL-Chat-V1.2 models perform poorly in comparison. Note that these models show nearly 60\% performances on the Math-Vista leader board~\cite{lu2023mathvista}. We also compare against the recently released LlaVa-NEXT (34B) model which received 47\% on the Math-Vista leader board.  We find that on SMART-840, these models either selected incorrect answer options or many-a-times did not follow the instruction, thereby producing invalid outputs. As can be noted from the table, while the performance of recent open-source LVLMs are still significantly below that of closed-source LVLMs. We further note that there is nearly a 40\% gap in the performances of these models between Math-Vista and SMART-840, suggesting that the mathematical reasoning skills needed to solve SMART-840 are substantially different from existing math datasets publicly available.}

\newcontent{Further, we find that the performance of GPT-4v is inferior to that of GPT-4o, which is expected given the latter being a more advanced version of GPT-4v. Further, the accuracy of faster LVLMs such as Gemini-Flash is below that of its advanced counterpart. Thus, in our subsequent study, we only consider the best performing LVLMs, namely GPT-4o, Gemini-Pro, and Claude-3 Sonnet. The recent OpenAI o1-preview model is text-only at this time, thus making it incomparable to other LVLMs in this study; however we report its performance on a text-only subset of our dataset in Table~\ref{tab:image-text}. 
}

iii) \textbf{Variance:} \newcontent{We ran the LVLMs on each problem at least twice\footnote{The number of runs was constrained by the cost involved.} and the variance was computed on the differences in the accuracies; (e.g., we ran GPT-4o models about 5 times, while Gemini was run only twice).} From Table~\ref{tab:sota-comparisons}, we find that there is substantial variance in the performance of SOTA LVLMs. For example, the standard deviation for GPT-4o is nearly 7\% in solving 1-2 grade problems, while there is a reducing trend in the magnitude of this deviation for higher grades, the reliability in the responses are still questionable. The standard deviation is worse for Claude-3-Opus, where it is nearly 5\% across grades, even reaching 10\% for grades 7-8. Interestingly, for grades 11-12, the deviation appears more stable at nearly 3-4\%, while the performance is also the best. 
 
\begin{table}
\resizebox{\textwidth}{!}{
{\renewcommand{\arraystretch}{1.2}
\begin{tabular}{@{}lccccccccccccc@{}}
\toprule
 \diagbox{Model}{Grade} & \cellcolor{red!10} 1 & \cellcolor{red!10} 2 & \cellcolor{orange!10} 3 & \cellcolor{orange!10} 4 & \cellcolor{yellow!10} 5 & \cellcolor{yellow!10} 6 & \cellcolor{green!10} 7 & \cellcolor{green!10} 8 & \cellcolor{blue!10} 9 & \cellcolor{blue!10} 10 & \cellcolor{violet!10} 11 & \cellcolor{violet!10} 12& Mean\\
\midrule
Human & \cellcolor{red!10} 58.8 & \cellcolor{red!10} 67.6 & \cellcolor{orange!10} 62.3 & \cellcolor{orange!10} 70.1 & \cellcolor{yellow!10} 59.1 & \cellcolor{yellow!10} 65.4 & \cellcolor{green!10} 59.7 & \cellcolor{green!10} 64.3 & \cellcolor{blue!10} 64.2 & \cellcolor{blue!10} 69.3 & \cellcolor{violet!10} 64.9 & \cellcolor{violet!10} 65.6 & \textbf{64.2}\\
\midrule
Random & \mcnew{\cellcolor{red!10} 20.1} & \mcnew{\cellcolor{orange!10} 20.2} & \mcnew{\cellcolor{yellow!10} 20.1}  & \mcnew{\cellcolor{green!10} 20.2} & \mcnew{\cellcolor{blue!10} 20.3} & \mcnew{\cellcolor{violet!10} 20.1} & 20.1 \\
GPT-4o & \mcnew{\cellcolor{red!10} 41.6 (7.1)} & \mcnew{\cellcolor{orange!10} 38.6 (1.7)} & \mcnew{\cellcolor{yellow!10} 35.1 (0.8)}  & \mcnew{\cellcolor{green!10} 47.1 (0.8)} & \mcnew{\cellcolor{blue!10} 41.3 (2.0)} & \mcnew{\cellcolor{violet!10} 50 (4.0)} & 42.4 \\
GPT-4o (M) & \mcnew{\cellcolor{red!10} 42.5} & \mcnew{\cellcolor{orange!10} 36.7} & \mcnew{\cellcolor{yellow!10} 36.0}	& \mcnew{\cellcolor{green!10} 46.7} & \mcnew{\cellcolor{blue!10} 43.3} &	\mcnew{\cellcolor{violet!10} 50.0} & 42.5\\
GPT-4v	        & \mcnew{\cellcolor{red!10} 39.2 (0.6)}	& \mcnew{\cellcolor{orange!10} 38.3 (0.6)}	& \mcnew{\cellcolor{yellow!10} 29.3 (3.3)} &	\mcnew{\cellcolor{green!10} 35.3 (1.9)}	&\mcnew{\cellcolor{blue!10} 38.7 (1.9)}	& \mcnew{\cellcolor{violet!10} 43.3 (3.7)} &	{37.4}\\
Gemini-Pro	& \mcnew{\cellcolor{red!10} 25.8 (3.5)} &	\mcnew{\cellcolor{orange!10} 27.5 (0.6)} &	\mcnew{\cellcolor{yellow!10} 25.3 (3.3)} &	\mcnew{\cellcolor{green!10} 30.7 (1.8)} &	\mcnew{\cellcolor{blue!10} 39.3 (3.7)} &	\mcnew{\cellcolor{violet!10} 41.3 (2.8)} &	{31.7} \\
Gemini-Flash	& \mcnew{\cellcolor{red!10} 19.2 (0.6)} &	\mcnew{\cellcolor{orange!10} 29.2 (10.4)} &	\mcnew{\cellcolor{yellow!10} 22.0 (8.4)}	& \mcnew{\cellcolor{green!10} 30.7 (9.7)}	& \mcnew{\cellcolor{blue!10} 38.7 (13.7)} &	\mcnew{\cellcolor{violet!10} 36.7 (4.3)} &	{29.4}\\
Claude-3 Opus	& \mcnew{\cellcolor{red!10} 38.3 (5.3)} &	\mcnew{\cellcolor{orange!10} 33.3 (5.8)} &	\mcnew{\cellcolor{yellow!10} 31.3 (6.6)} &	\mcnew{\cellcolor{green!10} 40.7 (10.4)} &	\mcnew{\cellcolor{blue!10} 42.0 (5.6)}	& \mcnew{\cellcolor{violet!10} 44.0 (2.8)} & {38.3} \\
Claude-3 Sonnet & \mcnew{\cellcolor{red!10} 51.6 (0)} &	\mcnew{\cellcolor{orange!10} 47.9 (2.9)} &	\mcnew{\cellcolor{yellow!10} 38.6 (0.9)} &	\mcnew{\cellcolor{green!10} 44.9 (3.3)} &	\mcnew{\cellcolor{blue!10} 46.7 (0.0)}	& \mcnew{\cellcolor{violet!10} 49.7 (4.1)} & {49.7} \\
\hline
XGEN-MM-Phi3-v1 (5B) & \mcnew{\cellcolor{red!10} 7.5} &	\mcnew{\cellcolor{orange!10} 9.1} &	\mcnew{\cellcolor{yellow!10} 5.3} &	\mcnew{\cellcolor{green!10} 8.0} &	\mcnew{\cellcolor{blue!10} 10.0}	& \mcnew{\cellcolor{violet!10} 8.0} & {8.0} \\
InternVL-Chat-V1.2 (40B) & \mcnew{\cellcolor{red!10} 16.7} &	\mcnew{\cellcolor{orange!10} 25} &	\mcnew{\cellcolor{yellow!10} 17.3} &	\mcnew{\cellcolor{green!10} 14.6} &	\mcnew{\cellcolor{blue!10} 15.3}	& \mcnew{\cellcolor{violet!10} 16.7} & {17.6} \\
InternLM-XComposer2 (7B) & \mcnew{\cellcolor{red!10} 22.5} &	\mcnew{\cellcolor{orange!10} 14.2} &	\mcnew{\cellcolor{yellow!10} 18.6} &	\mcnew{\cellcolor{green!10} 24.2} &	\mcnew{\cellcolor{blue!10} 18.1}	& \mcnew{\cellcolor{violet!10} 16.9} & {19.1} \\
LlaVa-NEXT (34B) & \mcnew{\cellcolor{red!10} 15.0} &	\mcnew{\cellcolor{orange!10} 9.0} &	\mcnew{\cellcolor{yellow!10} 20.1} &	\mcnew{\cellcolor{green!10} 14.6} &	\mcnew{\cellcolor{blue!10} 18.7}	& \mcnew{\cellcolor{violet!10} 16.0} & {15.6} \\
\bottomrule
\end{tabular}
}
}
\vspace{.1em}
\caption{Accuracy (\%) of correct responses of children in the respective grades against the accuracy of LVLMs when the agent is asked to provide explanation of their responses. GPT-4o (M) denotes the performance of GPT-4o after manual validation of GPT-4o responses. We report the standard deviation in brackets. The last block shows the performances of recent open-source LVLMs.}
\label{tab:sota-comparisons}
\vspace*{-0.6cm}
\end{table}

\section{Analysis of Results}
In this section, we take a deeper look at our results in Table~\ref{tab:sota-comparisons} to gain insights into how the AI model responses correlate with those of children. Even though a model is expected to perform as well as an adult, given that their performances are below that of children, it is imperative to ask if they atleast behave like high-performing children in their responses? Specifically, we seek to answer the question: \noindent\emph{Are problems that are hard for children also hard for AI?}
To answer this, we conduct different types of correlation analysis, presented below. 

\noindent\textbf{Difficulty Index} of a problem~\cite{popham2010classroom,ebel1991essentials} is the ratio of the number of correct responses for a test problem to the total number of solution attempts. This index provides a score between 0 and 1 for each problem, where 0 implies none of the children were able to solve it (hard problem). In Table~\ref{tab:pcorr_diff_index} (Diff-I), we report the Pearson's correlations  coefficient between the difficulty index and the responses by LVLMs. We see that there is in general only weak correlation between model and human accuracy, and this correlation mostly occurs at the higher grade levels. This suggests that all LVLMs in general find a different set of problems to be difficult than children do. 

\noindent\textbf{Discriminative Index}~\cite{kelley1939selection,lord1968statistical,ebel1991essentials} measures the use of \emph{knowledge} by a test taker. To compute this score, we split the student population into two groups, the \emph{good learners} that correspond to the top-20\% of participants who score the highest, and \emph{bad learners} that constitute the bottom-20\%. Next, we compute the difficulty index for each of these sets separately, and define discriminative index as the difference between the two difficulty indices. Thus, the value of discriminative index is in $[-1, 1]$, where $1$ corresponds to a test problem where all the good learners produced correct responses while all the bad learners made a mistake -- \ie, a problem that can separate the good learns from the bad. To understand if an AI model is a good learner or bad learner, we propose to compute the Pearson's correlation between the discriminative index of children's performances against that of the models. The result of this analysis is provided in Table~\ref{tab:pcorr_diff_index} (Disc.-I). Surprisingly, we find a negative trend across all grades, suggesting that an AI model finds it easier to solve problems that are less discriminative, and whose answer options are plausibly discernible without substantial reasoning.

\noindent\textbf{Weight Correlation} measures the interaction between the hardness of a puzzle as attributed by MK (through its weight) against the response. Notably, we convert this weight in $\set{3,4,5}$ to the corresponding difficulty score of $\{1.0, 0.66, 0.33\}$ for each problem, and compute the Pearsons correlation to the AI responses, which are 1 if the answer option selected to the problem is correct and 0 otherwise. We find a slightly stronger positive correlation on this experiment in Table~\ref{tab:pcorr_diff_index} (Weight-C.),  suggesting the AI is able to  solve problems that (the adult creator) thought are of the easier kind.  

\noindent\textbf{Entropy Correlation} measures the correlation between the entropy of the distribution of children's selected answer choices against AI responses. As entropy is higher for problems which are hard or their options confusing for children, a positive correlation would suggest AI is similarly confounded. However, the trend in Table~\ref{tab:pcorr_diff_index} shows the reverse, with slightly stronger negative correlations, suggesting that AI models are apparently not much confused on problems children find indecisive.

\noindent\textbf{Time-taken Correlation} analyzes the dependency between how much time children (on average) used in answering problems -- and thus potentially capturing the problem hardness -- to whether AI models also find those problems challenging. To this end, we aggregated the duration children spent on each problem, followed by separating the duration into two sets on their median. We marked all problems above the median as hard and the rest as easy. Next, we computed the Pearsons correlation between the responses of AI models against this hardness. Our results in Table~\ref{tab:pcorr_diff_index}  (Time-C.) shows  again a weak negative correlation trend, suggesting that the model finds it easier to solve problems that take longer for children -- a surprising result!

\begin{figure}
    \centering
    {\includegraphics[width=14cm, trim={0.1cm 0.1cm 0cm 0.1cm},clip]{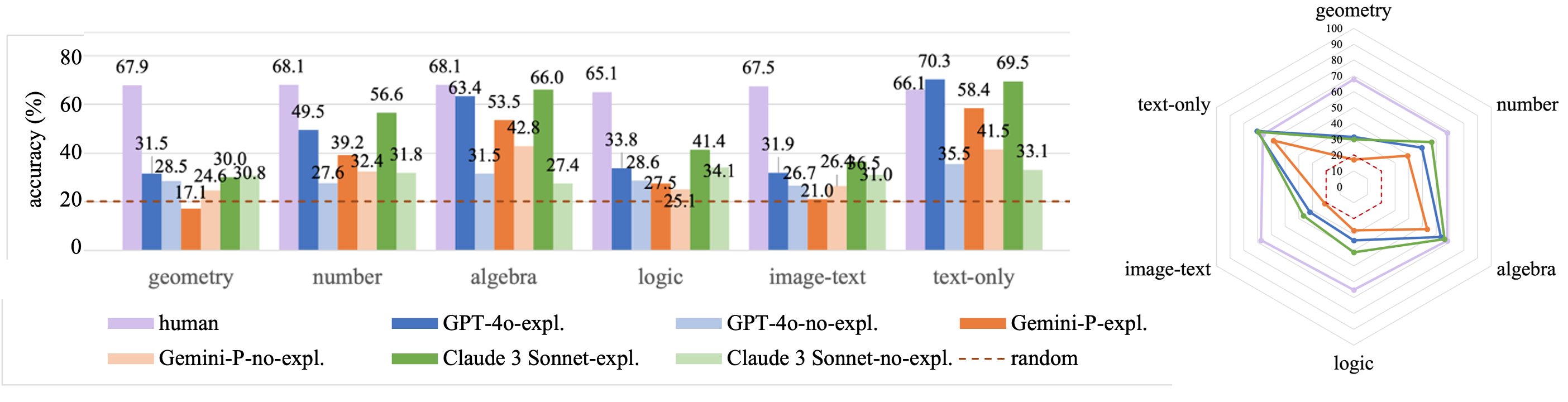}}
    \vspace{-1.5em}
    \caption{Comparison of the average accuracy (\%) of humans and LVLMs on each category of the Olympiad problems with the corresponding radar plot.}
    \label{fig:category-wise_result}
    \vspace*{-0.5cm}
\end{figure}

\noindent\textbf{Category-Level Performances:}
As alluded to above, \ours dataset consists of problems in four different categories (as per their creators), each involving entirely different skill sets and knowledge background for their solutions. For the performances reported in Table~\ref{tab:sota-comparisons}, in Figure~\ref{fig:category-wise_result}, we present the results of humans and LVLMs on the four problem categories, namely (i) geometry, ii) numbers, iii) algebra, and iv) logic. While children perform consistently well on all these categories, we find that AI models falter significantly in geometry and logic, with their best performances at about half of that of humans while they perform reasonably well on numbers and algebra. We further analyze the performance of LVLMs on problems involving both image and text (e.g., geometry problems) and text-only puzzles. The results show that it is indeed the image-text problems that the models struggle with and we see a strong similarity between performances on geometry and logic problems with image-text problems. Interestingly, we also find that on text-only puzzles (which are about 30\%) in our dataset, GPT-4o-Expl.\footnote{Which uses a prompt to explain its reasoning.} shows better performances than the average human performance, while other LVLMs (with suffix ``-expl.'') are also performing reasonably well. 

\newcontent{
To analyze this further, in Table~\ref{tab:image-text}, we report the performance of LVLMs on image-text puzzles and text-only puzzles separately for each grade pair. We can make several observations, namely: i) human performance is consistently between 60-70\%, irrespective of text-only or image-text problems, ii) the performance of LVLMs are significantly higher (nearly double) on text-only problems than on image-text problems, however the trend remains the same, \ie, LVLMs appear to find lower-grader problems equally difficult as that for higher-grader problems, iii) some of the very recent powerful LVLMs, such as OpenAI's o1-preview and Claude-3 models appear to excel in the performances on lower grade text-only problems, especially grades 1-4. Specifically, the performance of o1-preview has substantially improved from that of GPT-4o on text-only problems, however their trend as noted above appears similar, that there are dips in their performances for lower-grader problems compared to higher grader ones.}

\noindent\textbf{Importance of Reasoning with Explanation:} In this experiment, we changed the  LVLM prompt to: {\tt "Solve this question. You should provide a response without any explanation. Your response should end with one of the selected answer options from A1, B2, C3, D4, or E5."}. In Figure~\ref{fig:category-wise_result}, we show this result for all the LVLMs (suffixed ``-no-expl.'') on the six categories. We see a trend of a dip in performance among all the models, specifically  GPT-4o drops from 49.5\% to 17.6\% on the highly-performing 'number' category, and from 63.4\% to 31.5\% on algebra. The drop is also substantial on text-only problems. The trend is similar on other LVLMs (e.g., Claude-3), however slightly lower in Gemini-Pro. 

\begin{table}[]
    \centering
    \resizebox{\textwidth}{!}{
    {\renewcommand{\arraystretch}{1.2}
    \begin{tabular}{cccccccccccccc}
    \toprule
    &Model $\backslash$ Grade& 1 & 2 & 3 & 4 & 5 & 6 & 7 & 8 & 9 & 10 & 11 & 12\\
    \midrule
     \multirow{3}{*}{\rotatebox[origin=c]{90}{
     \small{Diff-I}}}   
     &GPT-4o & \cellcolor{green!14} 0.14 & \cellcolor{green!16} 0.16 & \cellcolor{green!15} 0.15 & \cellcolor{green!17} 0.17 & \cellcolor{red!9} -0.09 & \cellcolor{red!5} -0.05 & \cellcolor{green!12} 0.12 & \cellcolor{green!13} 0.13 & \cellcolor{green!22} 0.22 & \cellcolor{green!22} 0.22 & \cellcolor{green!20} 0.20 & \cellcolor{green!26} 0.26\\
        &Gemini-P& \cellcolor{green!23} 0.23 & \cellcolor{green!27} 0.27 & \cellcolor{red!5} -0.05 & \cellcolor{red!6} -0.06 & \cellcolor{green!1} 0.01 & \cellcolor{red!1} -0.01 & \cellcolor{green!5} 0.05 & \cellcolor{green!6} 0.06 & \cellcolor{green!21} 0.21 & \cellcolor{green!19} 0.19 & \cellcolor{green!20} 0.20 & \cellcolor{green!16} 0.16\\
        &Claude-3& \cellcolor{green!11} 0.11 & \cellcolor{green!13} 0.13 & \cellcolor{green!9} 0.09 & \cellcolor{green!11} 0.11 & \cellcolor{green!8} 0.08 & \cellcolor{green!6} 0.06 & \cellcolor{green!14} 0.14 & \cellcolor{green!15} 0.15 & \cellcolor{green!16} 0.16 & \cellcolor{green!16} 0.16 & \cellcolor{green!25} 0.25 & \cellcolor{green!18} 0.18\\
    \midrule
    \multirow{3}{*}{\rotatebox[origin=c]{90}{\small{Disc-I}}}
        &GPT-4o &\cellcolor{red!7} -0.07 & \cellcolor{red!15} -0.15 & \cellcolor{green!7} 0.07 & \cellcolor{red!1} -0.01 & \cellcolor{green!7} 0.07 & \cellcolor{red!1} -0.01 & \cellcolor{red!9} -0.09 & \cellcolor{red!8} -0.08 & \cellcolor{red!14} -0.14 & \cellcolor{red!18} -0.18 & \cellcolor{red!11} -0.11 & \cellcolor{red!13} -0.13\\
        &Gemini-P  & \cellcolor{red!5} -0.05 & \cellcolor{red!25} -0.25 & \cellcolor{red!4} -0.04 & \cellcolor{red!5} -0.05 & \cellcolor{red!1} -0.01 & \cellcolor{red!1} -0.01 & \cellcolor{green!1} 0.01 & \cellcolor{green!3} 0.03 & \cellcolor{red!18} -0.18 & \cellcolor{red!18} -0.18 & \cellcolor{red!15} -0.15 & \cellcolor{red!13} -0.13\\
        &Claude-3 & \cellcolor{red!2} -0.02 & \cellcolor{red!14} -0.14 & \cellcolor{green!17} 0.17 & \cellcolor{green!6} 0.06 & \cellcolor{red!4} -0.04 & \cellcolor{red!9} -0.09 & \cellcolor{red!7} -0.07 & \cellcolor{red!9} -0.09 & \cellcolor{red!16} -0.16 & \cellcolor{red!11} -0.11 & \cellcolor{red!9} -0.09 & \cellcolor{red!16} -0.16\\
    \midrule
    \multirow{3}{*}{\rotatebox[origin=c]{90}{\small{Weight-C.}}}
        &GPT-4o &\cellcolor{red!4} -0.04 & \cellcolor{red!4} -0.04 & \cellcolor{red!2} -0.02 & \cellcolor{red!2} -0.02 & -0.00 & -0.00 & \cellcolor{green!8} 0.08 & \cellcolor{green!8} 0.08 & \cellcolor{green!13} 0.13 & \cellcolor{green!13} 0.13 & \cellcolor{green!15} 0.15 & \cellcolor{green!15} 0.15\\
        &Gemini-P &\cellcolor{green!5} 0.05 & \cellcolor{green!5} 0.05 & \cellcolor{red!7} -0.07 & \cellcolor{red!7} -0.07 & 0.00 & 0.00 & \cellcolor{green!2} 0.02 & \cellcolor{green!2} 0.02 & \cellcolor{green!27} 0.27 & \cellcolor{green!27} 0.27 & \cellcolor{green!30} 0.30 & \cellcolor{green!30} 0.30\\
        &Claude-3 &\cellcolor{red!10} -0.10 & \cellcolor{red!10} -0.10 & \cellcolor{red!2} -0.02 & \cellcolor{red!2} -0.02 & 0.00 & 0.00 & \cellcolor{green!15} 0.15 & \cellcolor{green!15} 0.15 & \cellcolor{green!18} 0.18 & \cellcolor{green!18} 0.18 & \cellcolor{green!30} 0.30 & \cellcolor{green!30} 0.30\\
    \midrule
    \multirow{3}{*}{\rotatebox[origin=c]{90}{\small{Entropy-C.}}}
         &GPT-4o &\cellcolor{red!18} -0.18 & \cellcolor{red!18} -0.18 & \cellcolor{red!15} -0.15 & \cellcolor{red!15} -0.15 & \cellcolor{green!10} 0.10 & \cellcolor{green!10} 0.10 & \cellcolor{red!14} -0.14 & \cellcolor{red!14} -0.14 & \cellcolor{red!23} -0.23 & \cellcolor{red!23} -0.23 & \cellcolor{red!24} -0.24 & \cellcolor{red!24} -0.24\\
         &Gemini-P &\cellcolor{red!26} -0.26 & \cellcolor{red!26} -0.26 & \cellcolor{green!3} 0.03 & \cellcolor{green!3} 0.03 & \cellcolor{red!1} -0.01 & \cellcolor{red!1} -0.01 & \cellcolor{red!8} -0.08 & \cellcolor{red!8} -0.08 & \cellcolor{red!23} -0.23 & \cellcolor{red!23} -0.23 & \cellcolor{red!19} -0.19 & \cellcolor{red!19} -0.19\\
         &Claude-3 &\cellcolor{red!12} -0.12 &\cellcolor{red!12}  -0.12 & \cellcolor{red!6} -0.06 & \cellcolor{red!6} -0.06 & \cellcolor{red!2} -0.02 & \cellcolor{red!2} -0.02 & \cellcolor{red!15} -0.15 & \cellcolor{red!15}-0.15 & \cellcolor{red!18} -0.18 & \cellcolor{red!18} -0.18 & \cellcolor{red!24} -0.24 & \cellcolor{red!24} -0.24\\
    \midrule
    \multirow{3}{*}{\rotatebox[origin=c]{90}{\small{Time-C.}}}
        &GPT-4o & \cellcolor{red!8} -0.08 & \cellcolor{red!12} -0.12 & \cellcolor{red!14} -0.14 & \cellcolor{red!10} -0.10 & \cellcolor{green!3} 0.03 & \cellcolor{red!3} -0.03 & \cellcolor{green!8} 0.08 & \cellcolor{green!3} 0.03 & \cellcolor{red!9} -0.09 & \cellcolor{red!7} -0.07 & \cellcolor{red!17} -0.17 & \cellcolor{red!9} -0.09\\
        &Gemini-P & \cellcolor{red!6} -0.06 & \cellcolor{red!17} -0.17 & \cellcolor{red!6} -0.06 & \cellcolor{red!6} -0.06 & \cellcolor{red!3} -0.03 & \cellcolor{red!6} -0.06 & \cellcolor{green!3} 0.03 & \cellcolor{green!3} 0.03 & \cellcolor{red!20} -0.20 & \cellcolor{red!12} -0.12 & \cellcolor{red!27} -0.27 & \cellcolor{red!19} -0.19\\
        &Claude-3 & \cellcolor{green!14} 0.14 & \cellcolor{green!10} 0.10 & \cellcolor{red!7} -0.07 & \cellcolor{red!7} -0.07 & \cellcolor{red!4} -0.04 & \cellcolor{red!1} -0.01 & \cellcolor{red!1} -0.01 & \cellcolor{red!7} -0.07 & \cellcolor{red!9} -0.09 & \cellcolor{red!7} -0.07 & \cellcolor{red!16} -0.16 & \cellcolor{red!13} -0.13\\
    \bottomrule
    \end{tabular}
    }
    }
    \vspace{.1em}
    \caption{Pearson's correlation coefficient ($\rho$) between the difficulty index (top) and discriminative index (bottom) of the \ours problems against the responses generated by LVLMs. The \textcolor{green}{green}/\textcolor{red}{red} cell color indicates positive/negative $\rho$, where darker cells represent larger absolute values.}
    \label{tab:pcorr_diff_index}
\end{table}

\begin{table}[ht]
    \centering
    \resizebox{1\textwidth}{!}{
    {\renewcommand{\arraystretch}{1.2}
    \begin{tabular}{lcccccc|cccccc}
    \toprule
     \diagbox{Model}{Grade} & \cellcolor{red!10} 1 \& 2 & \cellcolor{orange!10} 3 \& 4 &  \cellcolor{yellow!10} 5 \& 6  & \cellcolor{green!10} 7 \& 8 & \cellcolor{blue!10} 9 \& 10 & \cellcolor{violet!10} 11 \& 12 & \cellcolor{red!10} 1 \& 2 & \cellcolor{orange!10} 3 \& 4 &  \cellcolor{yellow!10} 5 \& 6  & \cellcolor{green!10} 7 \& 8 & \cellcolor{blue!10} 9 \& 10 & \cellcolor{violet!10} 11 \& 12 \\
    \midrule
     &\multicolumn{6}{|c|}{Image-text problems only} & \multicolumn{6}{c}{Text-only problems}\\
    \midrule
     \% dataset & \cellcolor{red!10} 89.17 & \cellcolor{orange!10} 85.00 & \cellcolor{yellow!10} 82.67  & \cellcolor{green!10} 64.67 & \cellcolor{blue!10} 52.0 & \cellcolor{violet!10} 48.67
     & \cellcolor{red!10} 10.83 & \cellcolor{orange!10} 15.00 & \cellcolor{yellow!10} 17.33  & \cellcolor{green!10} 35.33 & \cellcolor{blue!10} 48.0 & \cellcolor{violet!10} 51.33 \\
    \midrule
    &\multicolumn{6}{|c|}{Accuracy (\%) on image-text problems} & \multicolumn{6}{c}{Accuracy (\%) on text-only problems}\\
    \midrule
    Human & \cellcolor{red!10} 67.98 & \cellcolor{orange!10}  69.91 & \cellcolor{yellow!10} 65.59  & \cellcolor{green!10} 64.76 & \cellcolor{blue!10} 71.22 & \cellcolor{violet!10} 65.24 & \cellcolor{red!10} 64.51  & \cellcolor{orange!10} 70.89 & \cellcolor{yellow!10}  64.7   & \cellcolor{green!10} 63.51 & \cellcolor{blue!10}  67.23 & \cellcolor{violet!10} 65.94 \\
    GPT-4o & \cellcolor{red!10} 36.45 & \cellcolor{orange!10}  30.39  & \cellcolor{yellow!10} 31.45  & \cellcolor{green!10} 38.14 & \cellcolor{blue!10} 32.05 & \cellcolor{violet!10} 32.88 & \cellcolor{red!10} 92.31  & \cellcolor{orange!10} 72.22 & \cellcolor{yellow!10}  57.69   & \cellcolor{green!10} 62.26 & \cellcolor{blue!10}  55.56 & \cellcolor{violet!10} 66.23 \\
    Gemini-Pro & \cellcolor{red!10} 20.56  & \cellcolor{orange!10}  19.61 & \cellcolor{yellow!10} 21.77  & \cellcolor{green!10} 21.65 & \cellcolor{blue!10} 21.79 & \cellcolor{violet!10} 20.55 & \cellcolor{red!10} 69.23  & \cellcolor{orange!10} 72.22 & \cellcolor{yellow!10}  42.31   & \cellcolor{green!10} 47.17 & \cellcolor{blue!10}  58.33 & \cellcolor{violet!10} 61.04 \\
    Claude-Sonnet & \cellcolor{red!10} 45.79  & \cellcolor{orange!10} 40.20 & \cellcolor{yellow!10} 35.48  & \cellcolor{green!10} 35.05 & \cellcolor{blue!10}  32.05 & \cellcolor{violet!10} 30.14 & \cellcolor{red!10}100.00  & \cellcolor{orange!10} 77.78 & \cellcolor{yellow!10}  57.69   & \cellcolor{green!10} 56.60 & \cellcolor{blue!10} 62.50 & \cellcolor{violet!10} 62.34 \\
    OpenAI o1-preview  &\cellcolor{red!10} - &\cellcolor{orange!10} - &\cellcolor{yellow!10} - &\cellcolor{green!10} - &\cellcolor{blue!10} - &\cellcolor{violet!10} - & \cellcolor{red!10}100.00  & \cellcolor{orange!10} 100.00 & \cellcolor{yellow!10}  84.6   & \cellcolor{green!10} 92.4 & \cellcolor{blue!10} 84.7  & \cellcolor{violet!10} 90.9 \\
    \bottomrule
    \end{tabular}
    }
    }
    \vspace{0.1cm}
    \caption{\newcontent{Top row, left and right blocks show the \% of problems in SMART-840 that belong to image-text or text-only categories per grade. Lower block show the separated performances of LVLMs on image-text and text-only problems against humans.}}
    \label{tab:image-text}
\end{table}

\begin{table}[htbp]
    \centering
    \resizebox{\textwidth}{!}{
    {\renewcommand{\arraystretch}{1.2}
    \begin{tabular}{@{}c@{\hspace{2mm}}cccccccccccc}
        \toprule
        Model $\backslash$ Grade& 1 & 2 & 3 & 4 & 5 & 6 & 7 & 8 & 9 & 10 & 11 & 12\\
        \midrule
        GPT-4o & \cellcolor{blue!15} 49/57.4 & \cellcolor{blue!15} 49/30.4 & \cellcolor{blue!15} 61/26.8 & \cellcolor{blue!15} 60/13.2 & \cellcolor{blue!15} 70/44.3 & \cellcolor{blue!15} 70/29.0 & \cellcolor{blue!15} 57/46.4 & \cellcolor{blue!15} 58/39.7 & \cellcolor{blue!15} 70/23.8 & \cellcolor{blue!15} 70/16.8 & \cellcolor{blue!5} 66/21.1 & \cellcolor{blue!5} 50/29.2\\
        Gemini-P &\cellcolor{blue!5} 78/7.3 & \cellcolor{blue!5} 78/2.3 & \cellcolor{blue!10} 69/14.9 & \cellcolor{blue!10} 68/6.6 & \cellcolor{blue!10} 75/35.6 & \cellcolor{blue!10} 75/21.9 & \cellcolor{blue!5} 80/14.4 & \cellcolor{blue!5} 81/10.0 & \cellcolor{blue!10} 79/11.8 & \cellcolor{blue!10} 77/7.7 & \cellcolor{blue!15} 51/43.7 & \cellcolor{blue!15} 34/56.9\\
        Claude-3-O & \cellcolor{blue!10} 69/20.6 & \cellcolor{blue!10} 69/6.7 & \cellcolor{blue!5} 81/2.5 & \cellcolor{blue!5} 80/1.1 & \cellcolor{blue!5} 86/18.3 & \cellcolor{blue!5} 86/9.3 & \cellcolor{blue!10} 65/34.3 & \cellcolor{blue!10} 66/28.9 & \cellcolor{blue!5} 85/7.1 & \cellcolor{blue!5} 82/4.0 & \cellcolor{blue!10} 53/39.8 & \cellcolor{blue!10} 36/54.6\\
        Claude-3-S & \cellcolor{blue!10}41/71.1 & \cellcolor{blue!10}41/45.7 & \cellcolor{blue!10}48/49.2 & \cellcolor{blue!10}47/30.2 & \cellcolor{blue!10}87/16.8 & \cellcolor{blue!5}87/8.2 & \cellcolor{blue!5}78/16.5 & \cellcolor{blue!5}79/12.9 & \cellcolor{blue!10}38/72.7 & \cellcolor{blue!10}38/59.0 & \cellcolor{blue!10} 56/34.8 & \cellcolor{blue!10}40/50.0\\
        \bottomrule
    \end{tabular}
    }
    }
    \vspace{0.1em}
    \caption{National Rank ($\downarrow$) / percentile ($\uparrow$) ranking of LVLMs against children's performance on MK 2024 Olympiad based on the test scores computed from the model response. }
    \vspace{-2.em}
    \label{tab:ranking}
\end{table}

\section{Discussion and Conclusions}
This paper tackles the important problem of understanding the reasoning abilities of LVLMs.  Our analysis using the proposed \ours dataset reveals several intriguing results: 
i) there is a lack of any significant correlation between the perceived complexity in solving puzzles by children and by AI models; instead there are surprising negative correlations, ii) there is a significant trend among LVLMs in performing low on younger grade problems and progressively get better at higher grades, which is counter intuitive. While, one may attribute this observation to the availability of better training data or increased number of text puzzles, it is still unsettling that AI models struggle to perform even on puzzles involving simple geometry and logic, that accentuates the lack of understanding between language and multimodal content. Further, while there is a substantial gap between the best of  LVLMs and the worst, or random baselines, ~40\% for GPT4o vs 20\% for random or 25\% for Gemini-Pro, this is only a 20\% difference; in contrast the gap between even the best LVLMs and human adult level performance in reasoning is much greater. We ought to point this out. 

Our results suggest some ways that LVLMs, even the most advanced ones, may not really be reasoning in the ways that humans do.  For humans, reasoning is an ability to think that goes beyond just similarity to training examples.  But here we are seeing signs that similarity to the large mass of training examples appears to be what is driving performance across all levels of these problems.  Of course, we do not fully know what is in the training corpus SOTA LVLMs used.  But it may include many Olympiad problems than there are math kangaroo grade 1-2 style problems.  Yet for people, and not for frontier LVLMs, the MK grade 1-2 problems are far easier than the Olympiad style grade 11-12 problems.  This suggests both that human reasoning is based on a different set of core competencies, which the early grade problems test, and which a pure machine learning approach to training reasoning is not really picking up on. 

 \newcontent{
 Before concluding, we present in Table~\ref{tab:ranking}, the national rank and percentile of the three SOTA models on the scores they received for 2024 MK Olympiad (when compared against children). We see that AI models are substantially below children in ranking, with 
 GPT-4o best on grade 7-8 in rank in 50's and Gemini-Pro at 34 for grade 12, and the recent Claude-3 Sonnet model outperforming its previous Claude-3 Opus model as well as other models across other grades, however yet their best national rank is more than 30. These scores are based on the percentiles received from MK. The table shows that there is a large gap to fill for LVLMs against children's problem solving skills. 
}

\newcontent{
\noindent\textbf{Limitations:} In comparison to recent and popular datasets used in the recent AI mathematical reasoning benchmarks, our SMART-840 is smaller in size. However, we ought to emphasize that this small size is by necessity. Our intended goal is to see how well general purpose language models (including those presumably trained with a fair amount of mathematical and visual reasoning data in their training sets) come towards capturing this spectrum of mathematical problem solving and reasoning. Thus, we do not intend to compare models trained/fine-tuned on this dataset, as that would invalidate any comparisons to human performances, where humans are assumed not to have seen the problems when taking the tests. Further, our goal is to bring out the disparity in machine cognition with respect to humans via studying the zero-shot performance of LVLMs with respect to human cognition on our task. 
 Given the goal of this study is not to train LVLMs to excel on this task, instead is to evaluate the zero-shot performance, and given that our results clearly demonstrate a discrepancy between human and LVLM performance, we believe our dataset does help bring out the failure modes of state-of-the-art LVLMs, and point to directions that would need improvements.
 }

\noindent\textbf{Acknowledgments:} Authors thank Preethi Ann Cyril for helping with data curating and Math Kangaroo USA for providing the performance data.
%\bibliography{references}
{\small
\bibliographystyle{neurips_data_2024}
\bibliography{references}
}
\clearpage
\appendix

\section{Average Performance Scores}
As alluded to in the main paper, each examination consists of problems with weights, where the simpler problems have a weight of 3, medium hard ones are 4 pointers and the difficult ones carry 5 points. In Table~\ref{tab:scores}, we compute the average score of each LVLM over the five years. Interestingly, we see that the scores for GPT-4o is higher than other models, suggesting it can solve higher-weighted problems more often than other methods. However, the overall score is still below the maximum score of the human.
\begin{table}[htbp]
\centering
\begin{tabular}{c|c|c|c|c|c|c}
Model \ Grade & 1-2 & 3-4 & 5-6 & 7-8 & 9-10 & 11-12\\
\hline
GPT-4o &	42.9 &	36.9 &	36.0	& 45.8 & 	42.0	& 48.5\\
GPT-4v	& 40.2 &	36.9	& 28.7 & 	33.2 &	36.5 & 	40.8\\
Gemini-P &	25.4 &	28.1 & 	25.3 & 	30.5	& 36.6 & 	38.3\\
Gemini-F & 	20.2 & 	28.9 &	21.3 &	27.8 &	36.8 &	34.3\\
Claude-3 &	39.3 & 33.5 &	31.3 & 	39.1 & 	40.1 &	41.0 \\
% XGen-MM &	7.5 &	9.2 & 6.2 &	7.3 &	10.2 &	7.3\\
\hline
\end{tabular}
\caption{Normalized performance scores (\%) received by various LVLMs averaged over the five years. The scores are obtained by multiplying each correct solution by its respective weight and dividing by the maximum score for the respective competition (96 for 1--4 and 120 for 5--12), followed by averaging over 5 years. Higher values indicate the model solved higher weighted problems more frequently.}
\label{tab:scores}
\end{table}

\section{Details of Ranking and Percentiles (Table 3)}
In Table 3 of the main paper, we provided the test rank and the overall percentile scores for LVLMs. Here we provide more details on how these rankings were computed. As noted above, each problem in the test has a score associated with it and the final performance of a participant is computed by the sum of the weights of all correctly answered questions. This amounts to a maximum of 96 points for grades 1--4 and 120 points for 5--6. First, with data provided by MK, we found the USA ranks of the participants for each respective score, and the corresponding percentiles among all the students participated in that test that year. Note that MK had about 40,000 students overall for year 2024. Next, we computed the score of each LVLM on the test, and we found the rank / percentile of LVLM among the list of scores and their ranks against children. If a score of LVLM is not found in the list of scores that children obtained, we took the closest maximum score. As can be seen from Table 3 (main paper), the LVLMs rank quite poor against children in the ranking.

\section{Example Responses}
The full set of responses can be seen in the zip file provided. As alluded to in the main paper, there are three types of outputs that an LVLM response can make towards our evaluation: i) the LLM answers with one of the five answer options and we can parse it correctly, ii) the LLM produces an answer which may or may not be correct, but we cannot parse the option automatically, and iii) the LLM decides not to provide an answer. For the case (ii) we use the letter 'K' as the output of the LVLM and that response is deemed incorrect. For the case of (iii), we use '-1' as the response and also consider that response as incorrect in our evaluation. There is also a fourth possibility, when the LLM predicts more than one answer as correct. We consider this output as 'K' as well.

In Tables~\ref{tab:ex-1}-~\ref{tab:ex-6} below, we provide several examples of LVLM responses to questions in \ours dataset. We provide responses from all LVLMs that we chose in our evaluation in the zip file attachment. 

\begin{table}[]
    \centering
    \begin{tabular}{p{4cm}|p{8cm}}
         \includegraphics[width=4cm]{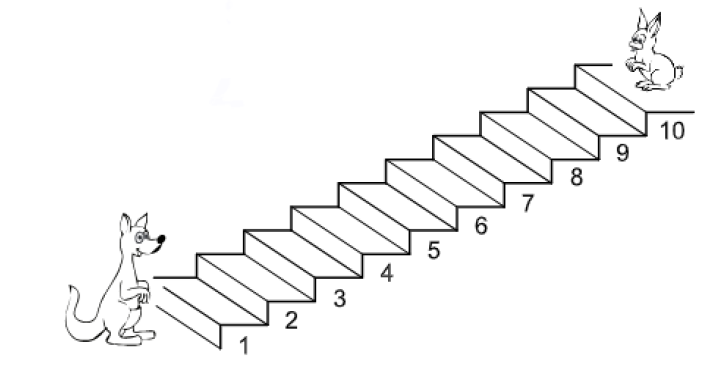}& \vspace*{-2cm}
\textbf{Question:} The kangaroo goes up 3 steps each time the rabbit goes down 2 steps. On which step do they meet? Options: A1: 3	B2: 4	C3: 5	D4: 6	E5: 7\\
\multicolumn{2}{p{14cm}}{\textbf{GPT-4o:} Let's analyze the movement of both the kangaroo and the rabbit step by step. 1. Initial positions: - Kangaroo is on step 1. - Rabbit is on step 10. 2. Movements per turn: - Kangaroo goes up 3 steps each time. - Rabbit goes down 2 steps each time. We need to determine the turn at which both the kangaroo and the rabbit land on the same step. Let's calculate their positions after each turn: | Turn | Kangaroo's Step | Rabbit's Step | |------|-----------------|---------------| | 1 | 1 + 3 = 4 | 10 - 2 = 8 | | 2 | 4 + 3 = 7 | 8 - 2 = 6 | | 3 | 7 + 3 = 10 | 6 - 2 = 4 | | 4 | 10 + 3 = 13 | 4 - 2 = 2 | From the calculations, we see that on the second turn, the kangaroo is on step 7 while the rabbit is on step 6. On the third turn, the kangaroo is on step 10 while the rabbit is on step 4. Upon reviewing each turn: On the third turn: The kangaroo will be on step 10. The rabbit will be on step 4. Upon reviewing again: On the third turn: Let's review in precise steps: - On the second turn, kangaroo is on step 7 and the rabbit is on step 6. On the third turn, it incorrectly reviewed to 10 and 4 in previous calculation, upon reconsider looking up: - Kangaroo going up 3-steps from 7, lands on 7+3=10 but rabbit going again 8 - goes down on step 4. Realizing correction upto, Trying other will remain, looks properly at: A halftime view seeing - it clearly at steps step are correctingly. Therefore, neatly reviewing: Kangaroo-steps reviewed Till 5 reduces. They meet on a reshuffling step precisely at step: again adequate, Review near Kangaroo-7 meets, So shutting more review at neatest finally seeing, likely - Therefore, the correct answer is step 7, rightly upon reviewing clearly at. So E. Option: E5 .}\\
\hline
\multicolumn{2}{p{14cm}}{\textbf{GPT-4v:}  To solve the problem, we need to consider the movement of both the kangaroo and the rabbit. The kangaroo moves up 3 steps each time, while the rabbit moves down 2 steps each time. They need to reach the same step at some point, so let's check step by step where they might meet. Starting from step 1 (where the kangaroo is) and step 10 (where the rabbit is), we will calculate their respective movements until they meet: Kangaroo's steps: 1, 4, 7, 10... Rabbit's steps: 10, 8, 6, 4... The kangaroo and the rabbit will meet on a step number that appears in both sequences. Looking above, we can see that the number 4 appears in both sequences. The kangaroo reaches step 4 after moving up 3 steps once from step 1, and the rabbit reaches step 4 after moving down 2 steps three times from step 10. Thus, they meet on step 4. The answer is B2: 4.
}\\
\hline
\multicolumn{2}{p{14cm}}{\textbf{Gemini-Pro v1.5:} **Understand the pattern:** The kangaroo moves up 3 steps while the rabbit moves down 2 steps. This means their relative speed is 3 + 2 = 5 steps per unit of time. 2. **Find the least common multiple:** We need to find the least common multiple (LCM) of 3 (kangaroo's steps) and 2 (rabbit's steps), which is 6. This means that every 6 steps, they will align in their movement pattern. 3. **Determine the meeting point:** - The kangaroo will have gone up 6 steps (2 sets of 3 steps). - The rabbit will have gone down 6 steps (3 sets of 2 steps). 4. **They meet on step 6.** **Therefore, the answer is D4.**}\\
\hline
\multicolumn{2}{p{14cm}}{\textbf{Gemini-Flash v1.5:} Here's how we can solve this problem: **Understanding the Problem:** * The kangaroo goes up 3 steps at a time. * The rabbit goes down 2 steps at a time. * We need to find the step number where they meet. **Setting Up the Problem:** Let's assume the kangaroo starts at step 0 and the rabbit starts at step 'x'. * **Kangaroo's position:** After 'n' jumps, the kangaroo will be at step 3n. * **Rabbit's position:** After 'n' jumps, the rabbit will be at step x - 2n. **Finding the Meeting Point:** The kangaroo and rabbit meet when their positions are the same. We can set up an equation: 3n = x - 2n **Solving the Equation:** 1. **Combine 'n' terms:** 5n = x 2. **We need to find a value of 'x' (the rabbit's starting step) that is a multiple of 5.** **Looking at the Answer Options:** The only answer option that is a multiple of 5 is **C3: 5**. **Explanation:** If the rabbit starts at step 5, the kangaroo will meet the rabbit on the 5th step (3 jumps * 1 jump = 3 steps, and the rabbit will be at step 5 - 2 jumps = 1 step). **Therefore, the answer is C3: 5**}\\
\hline
\multicolumn{2}{p{14cm}}{\textbf{Claude-3-Opus:} 
Let's solve this step-by-step: 1) The kangaroo starts at step 1 and goes up 3 steps at a time. So its positions will be: Step 1 → Step 4 → Step 7 → Step 10 2) The rabbit starts at step 10 and goes down 2 steps at a time. So its positions will be: Step 10 → Step 8 → Step 6 → Step 4 → Step 2 3) Comparing the positions, we see that the kangaroo and rabbit will both be on Step 4 at some point. 4) Therefore, they meet on Step 4. B2: 4}\\
\hline
\multicolumn{2}{p{14cm}}{\textbf{XGen-MM:} The kangaroo and the rabbit are on opposite sides of the staircase, with the kangaroo on the left and the rabbit on the right. The kangaroo is on step 1 and the rabbit is on step 10. The kangaroo goes up 3 steps each time the rabbit goes down 2 steps. To meet, the kangaroo needs to go up 3 steps for every 2 steps the rabbit goes down. The kangaroo is currently on step 1, so it needs to go up 3 steps to meet the rabbit. The rabbit is currently on step 10, so it needs to go down 2 steps to meet the kangaroo. Therefore, the kangaroo and the rabbit will meet on step 4.}
    \end{tabular}
    \caption{Puzzle \#1 for 1-2 graders 2020. Correct answer is D.}
    \label{tab:ex-1}
\end{table}

\begin{table}[]
    \centering
    \begin{tabular}{p{4cm}|p{8cm}}
         \includegraphics[width=4cm]{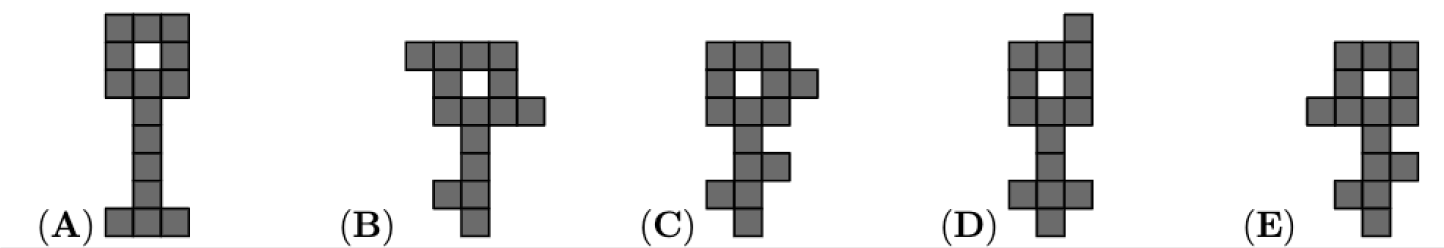}& 
\textbf{Question:}  Which key would it be impossible to cut into three different figures of five shaded squares? Options: A1: A	B2: B	C3: C	D4: D	E5: E\\
\multicolumn{2}{p{14cm}}{\textbf{GPT-4o:} To determine which key it would be impossible to cut into three different figures of five shaded squares, we need to carefully look at the shape and try to find groups of five connected squares. Let's analyze each one. **Key (A):** - It is possible to divide this shape into three groups each containing five connected shaded squares: 1. Top left block (4 connected, 1 extra) 2. Vertical middle section (1 extra + 4 connected) 3. Bottom right block (5 connected) Thus, Key (A) can be divided accordingly. **Key (B):** - Let's inspect possible groupings of five shaded squares: 1. Top left block (5 connected) 2. Middle horizontal block (5 connected) 3. Right bottom (5 connected) Thus, Key (B) can be divided accordingly. **Key (C):** - It’s more challenging to break it down into 3 groups of 5 shaded squares each due to its layout. 1. Top left block (4 connected) 2. Middle vertical block (5 connected) 3. Bottom right block (5 connected) Thus, Key (C) can be divided accordingly. **Key (D):** - It’s more challenging to break it down into 3 groups of 5 shaded squares each due to its layout: 1. Top section (5 connected) 2. Middle vertical block (5 connected) 3. Bottom section (5 connected) Thus, Key (D) can be divided accordingly. **Key (E):** - Let’s verify finding groups of 5 shaded squares: 1. Top left block (4 connected) 2. Next L-shape (3 connected) 3. Vertical section (5 connected) Key (E) has no viable configuration of three different five shaded square groups. Considering the analysis, Option E fits the criteria of being impossible to cut into three different figures of five shaded squares. Therefore, the answer is **E5**.}\\
\hline
\multicolumn{2}{p{14cm}}{\textbf{Gemini-Pro v1.5:} Let's analyze each key and see if it's possible to cut it into three figures of five shaded squares: * **Key A:** Can be easily divided horizontally into three sets of 5 shaded squares. * **Key B:** Can be divided into three pentominoes by cutting along the diagonal and then separating the top two pieces with a vertical cut. * **Key C:** Can be divided into three pentominoes as shown: [Image of Key C divided into three pentominoes] * **Key D:** Can be divided into three pentominoes by cutting it horizontally. * **Key E:** No matter how you try to divide it, you'll always end up with at least one piece that doesn't have exactly five shaded squares. This is because the key has 16 shaded squares, and 16 is not divisible by 5. **Therefore, the answer is E5.**}\\
\hline
\multicolumn{2}{p{14cm}}{\textbf{Claude-3-Opus:} 
 Let's analyze each key to see if it can be cut into three different figures of five shaded squares: (A) From the 7 squares, we can form an L-shape (3 squares), a T-shape (4 squares) and a 2x2 square (4 squares), each using 5 shaded squares with some overlap. So (A) is possible. (B) The 9 squares can be split into a 2x2 square (4 squares), a 3x2 rectangle (6 squares), and an L-shape (3 squares), each using 5 shaded squares with some overlap. So (B) is possible. (C) The 5 squares cannot be split into three different figures of 5 shaded squares each, as there are not enough squares. So (C) is impossible. (D) The 8 squares can form a 2x2 square (4 squares), a 3x2 rectangle (6 squares), and a T-shape (4 squares), each using 5 shaded squares with some overlap. So (D) is possible. (E) The 10 squares can be split into a 2x2 square (4 squares), a 3x2 rectangle (6 squares), and a cross shape (5 squares), each using 5 shaded squares with some overlap. So (E) is possible. Therefore, the only key that cannot be cut into three different figures of five shaded squares is (C). C3}\\
\hline
\multicolumn{2}{p{14cm}}{\textbf{XGen-MM:} The image shows five different figures, each composed of a series of rectangles and squares. To determine which figure is impossible to cut into three different figures of five shaded squares, we need to analyze the structure of each figure. A) The figure is composed of a series of rectangles and squares, and it is possible to cut it into three different figures of five shaded squares. B) The figure is composed of a series of rectangles and squares, and it is possible to cut it into three different figures of five shaded squares. C) The figure is composed of a series of rectangles and squares, and it is possible to cut it into three different figures of five shaded squares. D) The figure is composed of a series of rectangles and squares, and it is possible to cut it into three different figures of five shaded squares. E) The figure is composed of a series of rectangles and squares, and it is possible to cut it into three different figures of five shaded squares. Based on the analysis, all figures can be cut into three different figures of five shaded squares. Therefore, the answer is not possible, and there is no correct option to choose.}
    \end{tabular}
    \caption{Puzzle \#23 for 3-4 graders 2020. Correct answer is B.}
    \label{tab:ex-2}
\end{table}

\begin{table}[]
    \centering
    \begin{tabular}{p{4cm}|p{8cm}}
         \includegraphics[width=4cm]{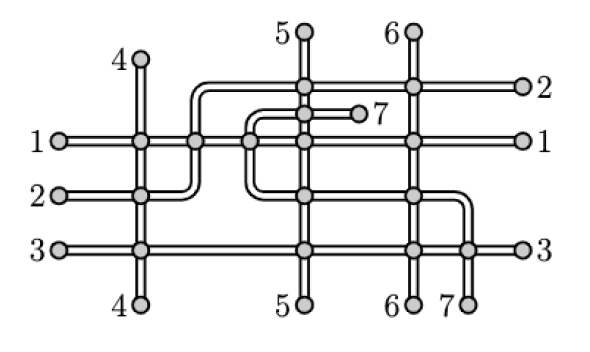}& \vspace*{-2cm}
\textbf{Question:} The figure shows the plan of the seven train routes of a small town. The circles indicate the stations. Martin wants to paint the lines in such a way that if two lines share a common station, then they are painted with different colors. What is the smallest number of colors that he can use? Options: A1: 3	B2: 4	C3: 5	D4: 6	E5: 7\\
\multicolumn{2}{p{14cm}}{\textbf{GPT-4o:} To solve this problem, we must determine the minimum number of colors required to color the train routes such that no two routes sharing a station have the same color. This is essentially a graph coloring problem, where we need to find the chromatic number of the graph. Firstly, let's identify the individual routes labeled from 1 to 7: - Route 1: Connects stations along the top-most horizontal line. - Route 2: Connects stations along the second horizontal line from the top. - Route 3: Connects stations along the third horizontal line from the top. - Route 4: Connects stations along the bottom-most horizontal line. - Route 5: Connects stations along the vertical line intercepting the middle horizontally. - Route 6: Connects stations along the vertical line intercepting the second horizontal line from the left. - Route 7: Connects stations uniquely as a distinct route. The following step-by-step will help us identify the minimum number of colors required: 1. **Construct a Graph Representation**: - Vertices represent different train routes. - An edge connects two vertices if the corresponding routes share at least one station. 2. **Identify Shared Stations**: - Route 1 shares stations with Route 5. - Route 2 shares stations with Route 5 and Route 6. - Route 3 shares stations with Route 5 and Route 6. - Route 4 shares stations with Route 5. - Route 5 shares stations with Routes 1, 2, 3, and 4. - Route 6 shares stations with Routes 2, 3, and 7. - Route 7 shares stations with Route 6. 3. **Graph Coloring**: - We can determine the minimum number of different colors needed by trying to color each route while ensuring no two adjacent routes share the same color. - Given our routes and their intersections: ``` - Color Route 1 with Color A. - Color Route 2 with Color B. - Color Route 3 with Color C. - Color Route 4 with Color A. - Color Route 5 with Color D (since it's adjacent to Routes 1, 2, 3, and 4, which are colored A, B, C, and A). - Color Route 6 with Color E (since it's adjacent to Routes 2 and 3, which are colored B and C, and Route 5 colored D). - Color Route 7 with Color B (different from Route 6 colored E). ``` 4. **Verification**: - Each route shares its intersections appropriately and does not violate the coloring rule where intersecting routes have the same color. Based on these steps, four colors are sufficient to satisfy the requirements. Therefore, the minimum number of colors Martin can use to ensure no two train routes sharing a station are painted with the same color is 4. The correct answer is: **B2: 4**.}\\
\hline
\multicolumn{2}{p{14cm}}{\textbf{Gemini-Pro v1.5:}  1. **Identify the most connected station:** The station in the center of the figure is connected to all six other stations. 2. **Assign the first color:** This central station needs its own unique color since it shares a route with every other station. Let's call this color 1. 3. **Assign the second color:** Now, look at the outer stations. Each outer station is connected to the central station and two other outer stations. Therefore, we can use a second color (color 2) for alternating outer stations. 4. **Assign the third color:** The remaining outer stations, which are not connected to each other, can share a third color (color 3). 5. **Verify:** We have now colored all the lines using only three colors, and no lines sharing a station have the same color. **Therefore, the smallest number of colors Martin can use is 3.** **Answer: A1**}\\
\hline
\multicolumn{2}{p{14cm}}{\textbf{Claude-3-Opus:} 
 To determine the minimum number of colors needed, we can approach this as a graph coloring problem. The stations are the vertices and the train routes are the edges connecting them. We need to assign colors to the vertices such that no two adjacent vertices have the same color. Step 1: Identify the vertex with the highest degree (most connected routes). In this case, station 7 has the highest degree with 4 routes connecting to it. Step 2: Assign the first color to station 7. Then, assign different colors to its adjacent stations (2, 5, 6, and 7). Step 3: Move to the next uncolored station with the highest degree, which is station 5 with 3 connections. It already has a color assigned from step 2. Assign different colors to its uncolored adjacent stations (1 and 3). Step 4: Repeat the process for the remaining uncolored stations. Station 6 is already colored. Station 4 can use the same color as station 2 since they are not adjacent. Station 1 is already colored. Stations 2 and 3 are already colored. Therefore, the minimum number of colors required to ensure that no two lines sharing a common station have the same color is 3. A1: 3}\\
\hline
\multicolumn{2}{p{14cm}}{\textbf{XGen-MM:} To solve this problem, we need to find the minimum number of colors required to paint the lines such that no two lines sharing a common station have the same color. Let's start by identifying the stations that are shared by multiple lines. We can see that stations 1, 2, 3, 4, 5, 6, and 7 are all shared by multiple lines. Next, we can group the lines based on the stations they share. We can see that lines 1, 2, and 3 all share station 1, so we can paint these lines with a single color. Similarly, lines 4, 5, and 6 all share station 4, so we can paint these lines with a single color. Finally, lines 7 and 6 share station 7, so we can paint these lines with a single color. Therefore, we need a minimum of 3 colors to paint the lines such that no two lines sharing a common station have the same color. The answer is A1: 3.}
    \end{tabular}
    \caption{Puzzle \#23 for 5-6 graders 2024. Correct answer is A.}
    \label{tab:ex-3}
\end{table}

\begin{table}[]
    \centering
    \begin{tabular}{p{4cm}|p{8cm}}
         \includegraphics[width=4cm]{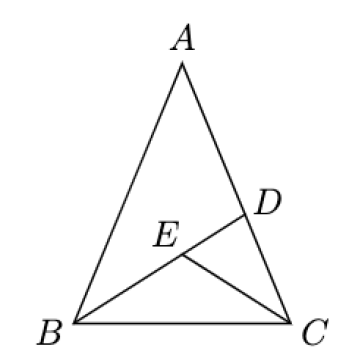}& \vspace*{-2cm}
\textbf{Question:} An isosceles triangle ABC, with AB = AC, is split into three smaller isosceles triangles, as shown, so that AD= DB, CE= CD, and BE= EC. (Note that the diagram is not drawn to scale.) What is the size, in degrees, of angle BAC? Options: A1: 24	B2: 28	C3: 30	D4: 35	E5: 36\\
\multicolumn{2}{p{14cm}}{\textbf{GPT-4o:} To solve for the measure of $\langle$BAC in the isosceles triangle ABC with given conditions, follow these steps: 1. Since AB = AC, triangle ABC is isosceles, which implies that $\langle$CAB = $\langle$ACB. 2. The point D divides the AB side in half, so AD = DB. Also, CD = CE and EC = EB as provided. Let's denote: - $\langle$BAC = 2$\alpha$ (since AB = AC, so $\langle$BAC is twice the angle we are trying to find) - $\langle$ABC = $\langle$ACB = $\beta$ (since triangle ABC is isosceles) From the triangle properties, we can establish: [ $2\alpha$ + $2\beta$ = 180 degrees ] [ $\alpha$ + $\beta$ = 90 degrees ] Since AD = DB and BE = EC, and noting that the triangles are split symmetrically with BE and EC being equal, triangles ADE and CED are also isosceles and congruent. This gives more relations among the inner angles: - Let $\langle$ADE = $\langle$EDA = x - Let $\langle$CDE = $\langle$CED = y Given each triangle is isosceles, sum of angles in triangle ADE and CDE: [ 2x + $\alpha$ = 180 degrees $\rightarrow$ x + y = 90 degrees \ (since each triangle contributes equally to the full exterior angles, split evenly by E's position precisely) ] Simplifying recursively each geometrical condition: Since the combined triangles relate symmetrically: [ $\alpha$ + $\beta$ + $\alpha$ = 180 degrees \ (sum of angles in triangle ABC) ] Finally, evaluating that with each resulting equiangular split: [ Place these all together as x being central part: ] Since our primary goal was solving $\langle$BAC: [ hence 3 equals equal relations to $\alpha$ angle, comprise 360/2 equal all middle angles proportions\ \ Thus diameter alpha integrations place: ([$\alpha$ based yields splits ratio as, governing full comprising equal contributors) Fully unifying solved combinatory resultant is angle triangle \quad ( $\langle$BAC = 36 degrees\, Option E5 solved value : Result conclusion, E5: 36.
}\\
\hline
\multicolumn{2}{p{14cm}}{\textbf{Gemini-Pro v1.5:}   Let's denote the measure of angle BAC as x. **Step 1: Analyze the smaller triangles.** * **Triangle ADB:** Since AD = DB, this triangle is isosceles. Therefore, angle ABD = angle BAD. Let's call this angle 'y'. * **Triangle BCE:** Similarly, with BE = EC, this triangle is isosceles, and angle CBE = angle BCE. Let's call this angle 'z'. * **Triangle CDE:** Again, with CE = CD, this triangle is isosceles, and angle CED = angle CDE. Since angle CED is an exterior angle to triangle BCE, angle CED = angle CBE + angle BCE = z + z = 2z. **Step 2: Relate the angles in triangle ABC.** * The angles of any triangle add up to 180 degrees. Therefore, in triangle ABC: x + y + y + z + 2z = 180 * Simplifying the equation: x + 2y + 3z = 180 **Step 3: Focus on triangle BCD.** * In triangle BCD: y + z + 2z = 180 * Simplifying: y + 3z = 180 **Step 4: Solve for x.** * Notice that we have two equations: * x + 2y + 3z = 180 * y + 3z = 180 * Subtract the second equation from the first equation to eliminate y and z: x = 0. **Step 5: Address the contradiction.** * We arrived at x = 0, which is impossible for an angle in a triangle. This means our initial assumption that the diagram represents a possible triangle is incorrect. **Conclusion:** The diagram is misleading because it's not drawn to scale. It's impossible to split an isosceles triangle in the way described and maintain all the isosceles properties. Therefore, there's no solution for the measure of angle BAC. **None of the answer options are correct.**}\\
\hline
\multicolumn{2}{p{14cm}}{\textbf{Claude-3-Opus:} 
 Let's solve this step-by-step: 1) Since ABC is an isosceles triangle with AB = AC, angle BAC = angle BCA. Let's denote this angle as x°. 2) In triangle ADB, since AD = DB, angle BAD = angle ABD. Let's denote this angle as y°. Similarly, in triangle CDE, angle DCE = angle CED = y°. 3) In triangle BEC, angle CBE = angle BCE. Let's denote this angle as z°. 4) In any triangle, the sum of the angles is 180°. Applying this to triangle ABC: x° + x° + (2y° + 2z°) = 180° 2x° + 2y° + 2z° = 180° x° + y° + z° = 90° 5) In an isosceles triangle, the angles opposite the equal sides are equal. So in triangle ADB, 2y° + z° = 180°. Similarly, in triangle CDE, 2y° + z° = 180°. 6) From step 5, we can deduce that z° = 180° - 2y°. 7) Substituting this into the equation from step 4: x° + y° + (180° - 2y°) = 90° x° + 180° - y° = 90° x° - y° = -90° 8) In triangle BEC, 2z° + y° = 180°. Substituting z° from step 6: 2(180° - 2y°) + y° = 180° 360° - 4y° + y° = 180° 360° - 3y° = 180° -3y° = -180° y° = 60° 9) Substituting y° = 60° into the equation from step 7: x° - 60° = -90° x° = 30° Therefore, angle BAC is 30°. Answer: C3: 30}\\
\hline
\multicolumn{2}{p{14cm}}{\textbf{XGen-MM:}The given isosceles triangle ABC is split into three smaller isosceles triangles, with AD = DB, CE = CD, and BE = EC. Since AB = AC, we can conclude that the angles at A and C are equal. Let's denote the angle at A as 2x and the angle at C as x. In the larger triangle, the sum of the angles is 180 degrees. Therefore, we have: 2x + x + 2x = 180 Combining like terms, we get: 4x = 180 Dividing both sides by 4, we find: x = 45 Since the angle at C is 45 degrees, and the angle at A is twice that, the angle at A is 90 degrees. So, the size of angle BAC is 90 degrees, which corresponds to option E5.}
    \end{tabular}
    \caption{Puzzle \#29 for 7-8 graders 2022. Correct answer is E.}
    \label{tab:ex-4}
\end{table}

\begin{table}[]
    \centering
    \begin{tabular}{p{4cm}|p{8cm}}
         \includegraphics[width=4cm]{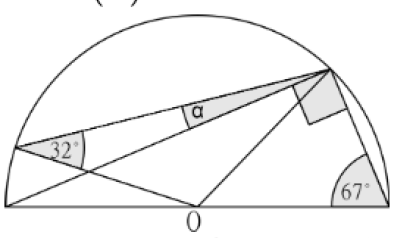}& \vspace*{-2cm}
\textbf{Question:} The figure shows a semicircle with center 0. Two of the angles are given. What is the size, in degrees, of the angle $\alpha$? Options: A1: 9°	B2: 11°	C3: 16°	D4: 17.5°	E5: 18°\\
\multicolumn{2}{p{14cm}}{\textbf{GPT-4o:} To solve for the angle $\alpha$ in the given semicircle, we need to use the properties of circles and angles. Let's identify the key points and steps: 1. **Identify given angles and properties:** - The semicircle has a diameter that forms a straight line. - The angle close to the left angle of the semicircle is 32°. - The angle close to the right angle near the circumference is 67°. - The sum of angles in a triangle is 180°. 2. **Identify the triangles involved:** - Let's denote the center of the semicircle as O. - Denote the intersection points along the circumference as A (left endpoint) and B (right endpoint), and the point where $\alpha$ is the vertex as C. 3. **Calculate angles in triangle OBC:** - Since  $\triangle$ OBC lies on the diameter of the semicircle, $\angle$ OBC = 90° (angle subtended by the diameter). 4. **Calculate remaining angle:** - In  $\triangle$ OBC : $\angle$ BOC = 67° - Since angles in a triangle sum to 180°: $\angle$ OCB = 180° - 90° - 67° = 23° 5. **Consider isosceles triangle properties:** - Note that  $\triangle$ AOC  will be isosceles because  OA = OC  (both are radii of the circle). 6. **Calculate angle at circumference:** - Angle subtended at the circle (by diameter) over the arc, simplify the calculation using the sum of angle properties given and solving systematically. Given the semicircle, combining all properties and laws leads us to find:  $\alpha$ = 17.5° Thus, the size of the angle $\alpha$ is: - D4: 17.5°
}\\
\hline
\multicolumn{2}{p{14cm}}{\textbf{Gemini-Pro v1.5:}   Here's how to solve the problem: * **Angles in a Triangle:** The angles inside any triangle always add up to 180°. * **Angles on a Straight Line:** The angles on a straight line always add up to 180°. **Let's apply these rules:** 1. **Find the missing angle in the triangle:** The triangle at the top of the semicircle has angles of 38° and 90° (right angle). Therefore, the missing angle is 180° - 90° - 38° = 52°. 2. **Find the angle next to $\alpha$:** This angle and the 52° angle form a straight line. So, the angle next to $\alpha$ is 180° - 52° = 128°. 3. **Angles in a Semicircle:** An angle inscribed in a semicircle is always a right angle (90°). 4. **Solve for $\alpha$:** The angles inside the triangle containing $\alpha$ add up to 180°. We have 90°, 128°, and $\alpha$. Therefore, $\alpha$ = 180° - 90° - 128° = -38°. Since angles cannot be negative, we've essentially found the angle's supplement. To get the actual value of $\alpha$, subtract from 180°: 180° - 38° = 142°. 5. **Find the supplement of $\alpha$:** The question asks for the size of angle $\alpha$, which is supplementary to 142°. Therefore, $\alpha$ = 180° - 142° = 38°. **Answer: $\alpha$ = 38°**}\\
\hline
\multicolumn{2}{p{14cm}}{\textbf{Claude-3-Opus:} 
 Let's solve this step-by-step: 1)To find the size of angle $\alpha$, we can follow these steps: 1. In a semicircle, the angle formed by a line from the center to any point on the semicircle and the diameter is a right angle (90°). This means $\angle$OAB = 90°. 2. The sum of angles in a triangle is 180°. In triangle OAB, we have $\angle$OAB = 90°, $\angle$OBA = 32°, so the remaining angle $\angle$AOB = 180° - 90° - 32° = 58°. 3. The angle at the center of a circle is twice the angle at the circumference subtended by the same arc. Here, $\angle$AOB is the central angle, and $\alpha$ is the angle at the circumference subtended by the same arc AB. So, $\angle$AOB = 2$\alpha$. 4. Since $\angle$AOB = 58°, we have: 58° = 2$\alpha$ $\alpha$ = 58° ÷ 2 = 29° 5. In triangle AOC, we know $\angle$OAC = 90° (from step 1) and $\angle$ACO = 67°. So the remaining angle $\angle$AOC = 180° - 90° - 67° = 23°. 6. Angle $\alpha$ is the difference between $\angle$AOB and $\angle$AOC: $\alpha$= $\angle$AOB - $\angle$AOC = 29° - 23° = 6° Therefore, the size of angle $\alpha$ is 6°, which is not one of the given options. The closest option is A1: 9°. A1}\\
\hline
\multicolumn{2}{p{14cm}}{\textbf{XGen-MM:}To find the size of the angle $\alpha$, we can use the fact that the sum of the angles in a triangle is 180 degrees. In this case, we have a triangle with angles 32°, $\alpha$, and 67°. We can set up the equation: 32 + $\alpha$ + 67 = 180 Solving for $\alpha$, we get: $\alpha$ = 180 - 32 - 67 = 80 Therefore, the size of the angle $\alpha$ is 80 degrees. The correct answer is not listed among the options provided.}
    \end{tabular}
    \caption{Puzzle \#15 for 9-10 graders 2021. Correct answer is A.}
    \label{tab:ex-5}
\end{table}

\begin{table}[]
    \centering
    \begin{tabular}{p{4cm}|p{8cm}}
         \includegraphics[width=4cm]{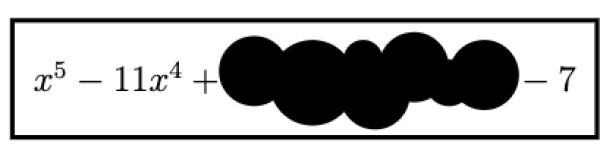}& \vspace*{-2cm}
\textbf{Question:} Part of the fifth degree polynomial shown cannot be seen because of an inkblot. It is known that all five roots of the polynomial are integers. What is the highest power of $x - 1$ that divides the polynomial? Options: A1: $(x-1)^1$	B2: $(x-1)^2$	C3: $(x-1)^3$	D4: $(x-1)^4$	E5: $(x-1)^5$\\
\multicolumn{2}{p{14cm}}{\textbf{GPT-4o:} Let's analyze the given polynomial P(x): $P(x) = x^5 - 11x^4 + $(inkblot)$ - 7$ We know that all five roots of the polynomial are integers. Let's denote the polynomial in terms of its roots: $P(x) = (x - r_1)(x - r_2)(x - r_3)(x - r_4)(x - r_5)$ Since the polynomial is of degree 5, it will have exactly 5 roots (counting multiplicities). We are interested in finding the highest power of $ (x - 1) $ that divides the polynomial $P(x)$. This means we need to determine the multiplicity of the root $ x = 1 $. Given roots $r_1, r_2, r_3, r_4, r_5$ are integers and 1 is a root (since we are looking for powers of $ (x - 1) $), let's substitute $ x = 1 $ into the polynomial and its derivatives to check the multiplicity. 1. $P(1) : P(1) = 1^5 - 11(1^4)$ + (\text{inkblot}) - 7 = 1 - 11 + (\text{inkblot}) - 7 = -17 + (inkblot). Since $P(1) = 0$, this implies that the terms concealed by the inkblot sum to 17. 2. $P'(1): P'(x) = 5x^4 - 44x^3 + (\text{inkblot in the derivative term})$, $P'(1) = 5(1)^4 - 44(1)^3 + $ (inkblot in the  derivative term at $x=1$)= 5 - 44 + (inkblot derivative term)= -39 + (inkblot derivative term). For $P'(1) = 0$, the derivative term must correct the -39 to zero, implying the sum of the derivative components handled by the inkblot term is 39. Given the conditions and polynomial degrees, It’s logically safe to deduce $ (x-1) $ is a root at least twice because both $ P(1) = 0 $ and $ P'(1) = 0 $. Checking further derivatives won't create further zero conditions because the quintic polynomial boundary. Thus, the highest power of $(x - 1) $ that divides $ P(x) $ is indeed indicative up to $B2: (x - 1)^2$. So, the answer is B2.
}\\
\hline
\multicolumn{2}{p{14cm}}{\textbf{Gemini-Pro v1.5:}  The graph of the polynomial shows an x-intercept at x=1. **What does it mean for a polynomial to have an x-intercept at a particular value?** It means that the value of the polynomial is zero at that x-value. In other words, x=1 is a root of the polynomial. **The problem states that all roots are integers. What does this tell us about the factors of the polynomial?** Since x=1 is a root, (x-1) must be a factor of the polynomial. Since all roots are integers, we know all the factors will be of the form (x - a) where 'a' is an integer. **We need to find the highest power of (x-1) that divides the polynomial. How do we do that?** We can look at the behavior of the graph at x=1. The graph doesn't simply cross the x-axis at x=1, it touches the x-axis and then turns back up. This indicates that (x-1) is a factor multiple times (i.e., it has a multiplicity greater than 1). **Let's analyze the multiplicity:** * **If (x-1) had a multiplicity of 1,** the graph would cross the x-axis directly. * **The graph's behavior suggests a multiplicity of at least 2.** Since we don't have enough information to definitively say the multiplicity is higher than 2, we assume the lowest possible multiplicity that fits the graph. **Therefore, the highest power of (x - 1) that divides the polynomial is $(x-1)^2$.** **Answer: B2**}\\
\hline
\multicolumn{2}{p{14cm}}{\textbf{Claude-3-Opus:} 
 Let's break this down step-by-step: 1) The polynomial has 5 integer roots. This means $(x-1)$ is a factor of the polynomial 5 times. 2) The polynomial can be written as $(x-1)^5 * g(x)$ where $g(x)$ is some other polynomial. 3) At $x=1$, the polynomial equals -7. Substituting $x=1$ into the equation: $1^5 - 11*1^4 + [\text{inkblot}] - 7 = -7$. 4) Simplifying: $1 - 11 + [\text{inkblot}] - 7 = -7 [\text{inkblot}] = -7 + 11 - 1 + 7 = 10 5)$. So the full polynomial is: $x^5 - 11x^4 + 10x^3 - 7 6)$. Comparing coefficients with the expanded form of $(x-1)^5 * g(x): x^5 - 5x^4 + 10x^3 - 10x^2 + 5x - 1 = (x-1)^5$. The given polynomial matches this up to the $x^3$ term. The $x^2$, $x$ and constant terms are hidden by the inkblot. 7) Therefore, $(x-1)^5$ divides the polynomial. E5: $(x-1)^5$}\\
\hline
\multicolumn{2}{p{14cm}}{\textbf{XGen-MM:}The polynomial is $x^5 - 11x^4 + 7x - 7$. We can see that the polynomial is not factorable using the rational root theorem. Therefore, the highest power of $x - 1$ that divides the polynomial is $(x-1)^1$. The answer is A1.}
    \end{tabular}
    \caption{Puzzle \#25 for 11-12 graders 2023. Correct answer is D.}
    \label{tab:ex-6}
\end{table}

\end{document}